\RequirePackage{fix-cm}
\documentclass{svjour3}                     
\smartqed  

\usepackage[utf8]{inputenc}
\usepackage{xcolor}
\usepackage{graphicx}
\usepackage{subfigure}
\usepackage{amssymb,amsmath}

\usepackage[normalem]{ulem}

\usepackage{tikz}
\usetikzlibrary{shapes,snakes}
\usepackage{ifthen}
\usepackage[]{algorithm2e}

\newcommand{\gras}[1]{\boldsymbol{#1}}
\newcommand{\mypar}[1]{\left(#1\right)}
\newcommand{\mya}[1]{\left\{#1\right\}}
\newcommand{\norme}[1]{\left\Vert #1\right\Vert_2}
\newcommand{\monabs}[1]{\left| #1\right|}

\newcommand{\Ephaz}{\mathcal{D}}
\newcommand{\Eobs}{\Omega}
\newcommand{\Esph}{\mathcal{S}}
\newcommand{\Eact}{\mathcal{A}}

\newcommand{\Nm}{n_p} 
\newcommand{\Nsnap}{N} 
\newcommand{\Na}	{n_{\action}}
\newcommand{\horvalue}	{K}	
\newcommand{\NULL}	{{\scriptscriptstyle \mathrm{NULL}}}
\newcommand{\NULLtext}	{{\footnotesize NULL}}

\newcommand{\fdyn}{\mathfrak{f}} 

\newcommand{\obs}{\gras{u}} 
\newcommand{\Point}{X} 
\newcommand{\bX}{\gras{X}} 

\newcommand{\Espoperator}{\mathbb{E}}

\newcommand{\monfloor}[1]{\left\lfloor #1\right\rfloor}

\newcommand{\obsdat}{y} 
\renewcommand{\obs}{\gras{y}} 
\newcommand{\Ne}{n_e} 
\newcommand{\Nv}{n_v} 
\newcommand{\ql}{w} 
\newcommand{\Nkey}{n_{\mathrm{key}}} 

\newcommand{\msys}{\mathcal{S}} 
\newcommand{\mep}{\mathcal{A}} 
\newcommand{\indA}{\mathcal{I}_\mep}
\newcommand{\indS}{\mathcal{I}_\msys}

\newcommand{\reward}	{R}
\newcommand{\cobj}	{\reward_{\mathrm{imm}}} 
\newcommand{\expvalue}{V^\pi}	

\newcommand{\drate}{\gamma}
\newcommand{\costfunction}{\mathcal{J}}

\newcommand{\ptrans}{p}
\newcommand{\action}{a}
\newcommand{\actionl}{a^{(l)}}
\newcommand{\policy}{\pi}

\newcommand{\state}	{s}	
\newcommand{\statei}	{s^{(i)}}	
\newcommand{\drag}	{{\mathcal{C}}}

\newcommand{\lagm}	{\rho} 

\newcommand{\effic}	{\eta}
\newcommand{\oracle}	{{\scriptscriptstyle \mathrm{ORA}}}
\newcommand{\oracletext}	{{\footnotesize ORA}}

\newcommand{\LMcomment}[1]	{}	
\newcommand{\Flocomment}[1]	{}	
\newcommand{\revision}[1]	{{\color{black}#1}}

\newcommand{\revLM}[1]	{}	

\begin{document}

\title{A statistical learning strategy for closed-loop control of fluid flows
}
\titlerunning{A statistical learning strategy for closed-loop control}


\author{Florimond Gu\'eniat        \and
        Lionel Mathelin \and
        M.~Yousuff Hussaini
}


\institute{
Florimond Gu\'eniat \at
Department of Mathematics,\\ Florida State Univ.,\\ 32306-4510 Tallahassee, FL, USA
\email{florimond@gueniat.fr}
\and
Lionel Mathelin \at
LIMSI-CNRS, rue J. von~Neumann,\\ Campus Universitaire d'Orsay,\\ 91405 Orsay cedex, France
\and
M. Yousuff Hussaini \at
Department of Mathematics,\\ Florida State Univ.,\\ 32306-4510 Tallahassee, FL, USA
}

\date{Received: date / Accepted: date}

\maketitle

\begin{abstract}
This work discusses a closed-loop control strategy for complex systems utilizing scarce and streaming data. A discrete embedding space is first built using hash functions applied to the sensor measurements from which a Markov process model is derived, approximating the complex system's dynamics. A control strategy is then learned using reinforcement learning once rewards relevant with respect to the control objective are identified. This method is \revision{designed for experimental configurations}, requiring no computations nor prior knowledge of the system, and enjoys intrinsic robustness. It is illustrated on two systems: the control of the transitions of a Lorenz 63 dynamical system, and the control of the drag of a cylinder flow. The method is shown to perform well.

\end{abstract}

\section{Introduction}
\label{sec-intro}
%
%
%
While the design and capability of aircraft, and more generally of complex systems, have significantly improved over the years, closed-loop control can bring further improvement in terms of performance and robustness to non-modeled perturbations. In the context of flow control, closed-loop control however suffers from severe limitations preventing its use in many situations.
As a paradigmatic example, a typical turbulent flow involves both a large range of spatial scales and exhibits a rich and fast dynamics. \revision{High frequency phenomena hence require a control command fast enough to adjust to the current state of the quickly evolving flow system. Indeed, frequencies of interest can routinely lie over 1 kHz, leading to a very short period of time for the controller to synthesize the command based on its knowledge of the state of the system.}


While flow manipulation and open-loop control are common practice, much fewer successful closed-loop control efforts are reported in the literature. Further, many of them rely on unrealistic assumptions. For example, Model Predictive Control (MPC) approaches require very significant computational resources to solve the governing equations in real-time. If a Reduced-Order Model (ROM) is employed, as is common practice to alleviate the CPU burden, one often needs to observe the whole system for deriving the ROM as, for instance, the  velocity or pressure fields with Proper Orthogonal Decomposition (POD), see \cite{Gerhard2003,Bergmann2008,Ma2011,Joe2011,Mathelin2012,Cordier2013}. Hence, flow control with this class of approaches is restricted to numerical simulations or experiments in a wind tunnel equipped with sophisticated visualization tools such as Particle Image Velocimetry (PIV).


This paper discusses a \emph{practical} strategy for closed-loop control of complex flows by alleviating the limitations of current methods.
The present work relies on a change of paradigm: we want to derive a general nonlinear closed-loop flow control methodology suitable for \emph{actual} configurations and as realistic as possible.
No \textit{a priori} model, nor even a model structure, describing the dynamics of the system is required to be available. The approach proposed is \emph{data-driven only}, with the sole information about the system given by scarce and spatially-constrained sensors. The method then exploits statistical learning methods.

This is the framework one typically deals with in practical situations where the amount of information on the system at hand is limited and usually comes from a few sensors located at the boundary of the fluid domain, \textit{e.g.}, on solid surfaces. The resulting information takes the form of short time-dependent vectors with as many entries as sensors. 

Among the few earlier efforts relying on streaming measurements from a few sensors, a trained neural network using surface measurements is employed to reduce the drag of a turbulent channel flow with an opposition control strategy in \cite{Lee1997}. In \cite{Kegerise2007}, pressure sensors and an auto-regressive model (AutoRegression with eXogenous inputs, ARX) are used to reduce flow-induced cavity tones. \revision{An autoregressive approach is also followed in \cite{Huang_Kim_2008} and \cite{Herve2012}, while a genetic programming technique is adopted in \cite{Gautier2015} to control a separated boundary layer.
Interested readers may refer to \cite{Brunton2015} for a comprehensive review of the topic. The present approach aims at deriving an efficient, yet robust, nonlinear closed-loop control method compliant with actual situations. Among its distinctive features compared to other methods is a combination of \emph{both} performance and fast learning.}

To facilitate \emph{learning} about the system dynamics from the time-dependent measurements, and the subsequent derivation of a control strategy, the problem must be amenable to a finite dimension. Hence, one needs to discretize the infinite-dimensional time-series of the sensors' information. To this end, the streaming data are convolved with a kernel which should result in a discrete image space. Locality-Sensitive Hash (LSH) functions are used for that purpose,~\cite{Slaney2008}, which results in a low-dimensional discrete state space. Transitions from state to state in this discrete space describe the dynamics,~\cite{Kaiser2014}, and allow the analysis to learn, and update in real-time, a Markov process model of the system.
A suitable discretization of the dynamics allows the derivation of a reinforcement learning-based control strategy of the identified Markov process model of the system. The control of Markov processes is a mature field,~\cite{Mandl1974} and reinforcement learning, \cite{Watkins1992,Gosavi2011}, is a suitable class of methods for the control of Markov processes, see for instance \cite{Lin1999,Gadaleta1999} for the control of 1-D and 2-D chaotic maps. As will be seen in the application examples below, the resulting control strategy is data-driven only, intrinsically robust against perturbations in the flow and does not require significant computational resources nor prior knowledge of the flow.
The proposed approach is experiment-oriented and on-going efforts are carried-out to demonstrate it on an experimental open cavity flow in a turbulent regime. This will be the subject of a subsequent publication.

The paper is organized as follows. The framework and basics of how hash functions are used to generate a low-dimensional state space are discussed in Section~\ref{sec-comp}. Section~\ref{Sec_Statedrivencontrol} is concerned with learning an efficient control strategy for the system modeled as a stochastic process living in a small dimensional space. The resulting control strategy is illustrated and discussed in the case of the control of a Lorenz 63 system and the drag reduction of a cylinder in a two-dimensional flow in Section~\ref{sec-results}. Concluding remarks close the paper in Section~\ref{sec-conclu}.

\section{Hash functions for reduced order modeling}
\label{sec-comp}

\subsection{Preliminaries}
Consider a dynamical system evolving on a manifold $\Ephaz$:
$$
\dot{\Point}\mypar{t} = \fdyn\mypar{\Point\mypar{t}}, \qquad \Point \in \Ephaz,
$$
with $\Point$ the state of the system and $\fdyn$ the flow operator. Let $\gras{g}:\Ephaz \rightarrow \mathbb{R}^{\Nm}$ be a sensor function. In the sequel, the number of sensors $\Nm$ will be taken to be one but generalization to more sensors is immediate. The observed data $\obsdat \in \mathbb{R}$ is defined as $\obsdat\mypar{t} := \gras{g}\mypar{\Point\mypar{t}}$.

Let $\Delta t$ be the sampling rate of the measurement system. Sampling has to be fast enough to capture the small time scales of the dynamics of $\obsdat$. The data coming from the sensor are embedded in a reconstructed phase space $\Eobs$:
$$\mypar{ \obsdat\mypar{t} \: \obsdat\mypar{t-\Delta t} \hdots \obsdat\mypar{t-\mypar{\Ne-1}\Delta t}} =: \obs\mypar{t} \in \Eobs \subset \mathbb{R}^{\Ne}.$$

The correlation dimension of $\obsdat$ is estimated from the time-series, for instance using the Grassberger-Proccacia algorithm, \cite{Grassberger1983}. It allows the definition of the embedding dimension $\Ne$ as, at least, twice the correlation dimension. Under mild assumptions, this resulting embedding dimension ensures there is a diffeomorphism between the phase space $\Ephaz$ and the reconstructed phase space, \cite{Takens1981}, so that $\obs$ is an observable on the system.

\subsection{Hash functions} \label{sec_hash_functions}
\revision{A hash function is any function that can be used to map an entry $\obs \in \mathbb{R}^{\Ne}$ to a \emph{key} $\state \in \mathbb{Z}$. Since the key is an integer, hash functions effectively result in a discrete image space of $\obs$.} Hash functions are often used to efficiently discriminate two different entries so that slightly different input data should result in a large variation of the associated key,~\cite{Carter1977}. An important objective in choosing the hash function is to avoid \emph{collisions}, \textit{i.e.}, when two different entries are associated with the same key. 

Conversely, the need for identification of similar entries in large databases has led to the use of the Locality-Sensitive Hash functions (LSH)~\cite{Andoni2006,Slaney2008}. In contrast with most hash functions, they are designed to promote collisions when two entries are close to each other. The idea is that, if two points are close in $\mathbb{R}^{\Ne}$, they should be likely to remain close after a projection on a lower dimensional subspace. The Johnson-Lindenstrauss lemma (JLL),~\cite{Johnson1984}, provides useful results to reach this objective and motivates the use of LSHs. \revision{Specifically, the JLL provides probabilistic guarantees of the near preservation of relative distances between objects in high-dimension after projection onto random low-dimensional spaces.}

Let $\gras{v} \in \mathbb{R}^{\Ne}$, $\left\|\gras{v}\right\|_2 = 1$, be a test vector. The function $h^{\gras{v},\ql}:\mathbb{R}^{\Ne} \rightarrow \mathbb{N}$ is a LSH, \cite{Andoni2006}:
\begin{equation}
 h^{\gras{v},\ql}\mypar{\obs} := h_0 + \monfloor{ \frac{\gras{v}\cdot \obs}{\ql}},
\label{eq-dhash}
\end{equation}
where $\monfloor{\cdot}$ is the floor operator, $\ql > 0$ a quantization length and $h_0$ is such that $h^{\gras{v},\ql} > 0$, $\forall \obs \in \Eobs$, $\gras{v} \in B_2^{\Ne}\left(1\right)$, with \revision{$B_2^{\Ne}\left(1\right)$ the unit ball of $\mathbb{R}^{\Ne}$ in the sense of $L^2$.}
\revision{As an illustration,} following ~\cite{Johnson1984,Slaney2008}, if two observables $\obs_1$ and $\obs_2$ are such that $\norme{\obs_1-\obs_2} < r$, with $r = w \slash 2 \norme{\gras{v}}$, they are associated with the same key with a probability $p_1$ larger than $1 \slash 2$. On the other hand, the probability $p_2$ that two distant points $\obs_1$ and $\obs_2$ appear close to each other in the sense of $h^{\gras{v},w}$ is a function of the angle between $\mypar{\obs_2 - \obs_1}$ and $\gras{v}$ and is smaller than $p_1$. 

\begin{figure}[t]
\center
\includegraphics[height=.4\linewidth]{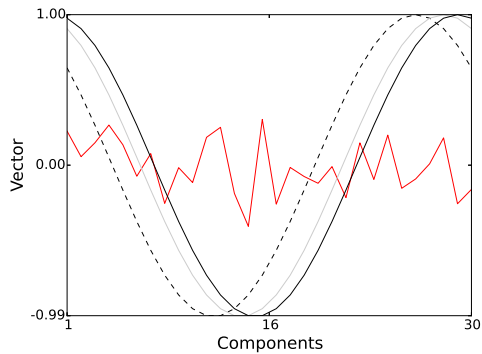}
\caption{\small \revision{Plot of the vectors $\gras{y}^a$ and $\gras{y}^b$ (solid lines) and $\gras{y}^c$ (dashed line). The random test vector $\gras{v}$ is plotted in red.}}
\label{fig-lsh}
\end{figure}

\revision{
Let us illustrate the LSH with a simple example.
%
%
%
Let $\Ne=30$ and $\gras{y}^a, \gras{y}^b$ and $\gras{y}^c$ be three vectors from $\mathbb{R}^{\Ne}$ such that:
$$
\begin{array}{ll}
y^a_j &= \cos\mypar{2 \, \pi \, \mypar{j-1}}, \qquad 1 \le j \le \Ne,\\
y^b_j &= \cos\mypar{2 \, \pi \, \mypar{j}},\\
y^c_j &= \cos\mypar{2 \, \pi \, \mypar{j+2}}.
\end{array}
$$
Let $\gras{v}$ be a unit-norm normal Gaussian vector of $\mathbb{R}^{\Ne}$, $\ql = 0.2$ and $h_0 = 0$.
The different vectors are drawn in Fig.~\ref{fig-lsh}. 
Upon processing with the LSH, the keys associated with the three vectors $\gras{y}^a, \gras{y}^b$ and $\gras{y}^c$ are:
$$
h^{\gras{v},\ql}\left(\gras{y}^a\right) = 3, \qquad h^{\gras{v},\ql}\left(\gras{y}^b\right) = 3, \qquad h^{\gras{v},\ql}\left(\gras{y}^c\right) = 1.
$$
$\gras{y}^a$ and $\gras{y}^b$ are closer to each other, in the sense of their $L^2$-distance, than to $\gras{y}^c$, and indeed, vectors $\gras{y}^a$ and $\gras{y}^b$ are associated with the same key $\state = 3$ while $\gras{y}^c$ is associated with $\state = 1$.
}

To discriminate false neighbors with higher probability, projections on several low-dimensional subspaces can be used.
Consider the hash function $\mathfrak{h}:\mathbb{R}^{\Ne} \rightarrow \Esph \subset \mathbb{N}$ made of concatenation of $\Nv$ keys $\left\{h^{\gras{v}_l,w_l}\mypar{\obs}\right\}_{l=1}^{\Nv}$.
Many choices can be made for the test vectors $\left\{\gras{v}_l\right\}_l$. For instance, they could be the principal axes of the manifold on which the observable $\obs$ evolves, or may be randomly selected from a Gaussian distribution. Keys (\emph{i.e.}, objects in the image space of $\mathfrak{h}$) generate a Vorono\"{i} paving of the observable space. Each key is associated with a cell, or \emph{state} $\state$, and close observables are associated with the same key.

Coarseness of the paving depends on the quantization lengths $\left\{\ql_l\right\}_l$ and the number $\Nv$ of test vectors.
The set $\left\{\ql_l\right\}_l$ quantifies the minimal length of the cells\footnote{Two vectors $\obs_1$ and $\obs_2$ are associated with two different keys if their $L^2$-norms differ by more than \revision{$\ql \slash \lvert \cos\left(\angle\left(\gras{v}, \obs_1 - \obs_2\right)\right) \rvert$,} hence the minimal length $\ql$ of the cells.} and, as $\ql_l$ increases, the cardinality $\Nkey := \mathrm{card}\left(\Esph\right)$ of the image space of $\mathfrak{h}$ decreases.
On the other hand, increasing $\Nv$ is equivalent to refining the description, \emph{i.e.}, increasing the cardinality of $\Esph$.

\section{State-driven control} \label{Sec_Statedrivencontrol}
Thanks to the hash function, the original infinite-dimensional system is approximated as a discrete stochastic process whose state space is spanned by the keys. The system is observable in real-time in this discrete space since evaluating the hash function with the streaming sensor data $\obs$ can be done at no computational cost. Under a discrete control command $\action \in \Eact$ to the actuators, hereafter termed the \emph{action}, the dynamics of the system will be modified and the goal is to find the action which makes the physical system at hand satisfy the control objective, say, a target dynamics.
The discretized description of the system with the hash function naturally lends itself to a Markovian framework,~\cite{Novikov1989,Renner2001}, which is adopted below.

\subsection{Markov Decision Process}

Let \revision{$\ptrans\mypar{i,l,j}$} be the probability of transition from a state \revision{$\statei \in \Esph$ to $\state^{(j)} \in \Esph$} under an action \revision{$\action^{(l)} \in \Eact$, $\Eact := \left\{\action^{(l)} \in \mathbb{R}, \, l \in \indA := \left[1, \Na\right] \subset \mathbb{N}\right\}$ being a uniformly discretized space of possible control actions. }Here again, the analysis is restricted to one actuator but generalization to more actuators is immediate.
Actions $\left\{\action^{(l)}\right\}_{l=1}^{\Na}$ index the discrete commands available to the controller. \revision{Define $\Esph := \left\{\statei, i \in \indS := \left[1, \Nkey\right] \subset \mathbb{N}\right\}$ and $\state_k \equiv \state\left(t_k := k \, \Delta t\right)$. Similarly, $\action_k \equiv \action\left(t_k\right)$.} To each transition-action $\mypar{i\rightarrow j, l}$ is associated a \emph{transition reward} (TR) \revision{$\reward\mypar{\statei,\actionl,\state^{(j)}}$} \footnote{\revision{We use a slight abuse of notation as we consider $\reward\mypar{\statei,\actionl,\state^{(j)}} \equiv \reward\mypar{i,l,j}$ and $p\mypar{i,\actionl,j} \equiv p\mypar{i,l,j}$. Both $\reward$ and $p$ are $\Nkey \times \Na \times \Nkey$ three-way tensors.}}. 
The goal of the control strategy is hence to identify the optimal \emph{policy} $\policy: \Esph \rightarrow \Eact$, which describes the best action to apply when in a given state so as to maximize the \emph{value}, \revision{$\expvalue_i$}, defined as the sum of the future expected rewards of the policy when starting at state $\statei$:
 \revision{\begin{equation}
  \expvalue_i := \lim_{\horvalue \rightarrow \infty} \Espoperator\left[\sum_{k=1}^\horvalue \drate^{k-1} \reward\mypar{\state_k,\policy\mypar{\state_k},\state_{k+1}} \right], \qquad \state_1 = \statei, \label{eq-value}
 \end{equation}}
where $\Espoperator \left[\cdot\right]$ is the expectation operator over all possible sequences $\left\{\state_k\right\}_{k=1}^\horvalue$ under the policy $\policy$. Here, $\drate$ is the discount rate, $0 < \drate < 1$. 
By weakly accounting for TRs occurring far in the future, the discounted rate effectively introduces a time horizon. 
The values express the expectation of the cumulative TRs of a given policy. 
Eq.~\eqref{eq-value} may be reformulated as, \cite{Mandl1974}:
\revision{\begin{equation}
 \expvalue_i =  R_i^{\policy} + \drate \, \sum_{j \in \Esph} p(i,\policy\mypar{\statei},j) \, \expvalue_j,
\label{eq-value_reform}
\end{equation}}
where $R_i^{\policy} := \Espoperator\left[\reward\left(\state_k = \statei, \policy\left(\state_k\right), \state_{k+1}\right)\right]$ is the mean TR in state \revision{$\state_k = \statei$ }under the policy $\policy$. This corresponds to the celebrated Bellman equation,~\cite{Bellman1952}, written in the discrete settings.


\subsection{Identification of the rewards}
The transition rewards are unknown \textit{a priori} and depend on the control objective. \revision{As perhaps more familiar to the reader, one could define the cost as the opposite of the reward. Consequently,} to learn relevant rewards, consider the cost function $\costfunction$ associated with the control objective:
\begin{equation}
\costfunction\mypar{t_n} := \sum_{k = 0}^\infty{\drate^k \, \left(\drag\mypar{t_{n+k+1}} + \lagm \, \monabs{\action\mypar{t_{n+k}}}^2 \right)},
\label{eq-costfunc}
\end{equation}
with $\drag$ the measure of performance, \textit{e.g.}, the drag coefficient in the included example. 
The contribution of the control intensity $\monabs{\action}^2$ to the cost with respect to the measure of performance $\drag$ is weighted by $\lagm > 0$.

Let the immediate rewards $\cobj\left(\state_n, \action_n, \state_{n+1}\right)$ represent, at any given time $t_n$, the negative of the contribution to the cost $\costfunction\mypar{t_n}$ of the transition from the present state $\state_n \equiv \state\left(t_n\right)$ to state $\state_{n+1}$, under an action $\action_n \equiv \action\left(t_n\right)$: 
\begin{equation}
\cobj\left(\state_n, \action_n, \state_{n+1}\right) := - \left(\drag\mypar{t_{n+1}} + \lagm \, \monabs{\action\mypar{t_{n}}}^2 \right)
\label{eq-inst_rew}.
\end{equation}

$\cobj$ is then high when the performance associated with the controlled system is good, and low otherwise.
\revision{Instead of the reward associated with a particular trajectory in the original, infinite-dimensional, phase space, the \emph{average}, trajectory-independent, reward associated with all trajectories leading to a given transition $\mypar{\state_n\rightarrow \state_{n+1},\action_n}$ should be determined. The ergodicity assumption postulates the equivalence of a temporal and an ensemble average, \textit{i.e.}, here $\displaystyle \lim_{\horvalue \rightarrow \infty} {\horvalue}^{-1} \, \sum_{k=0}^{\horvalue-1}{f\left(\state_k\right) = \Espoperator\left[f\left(\state\right)\right]}$. Under this assumption, the mean transition rewards, in the sense of the probability distribution of $\cobj$, are finally determined during the learning stage \textit{via}:}
\begin{equation}
\reward\left(\state_n, \action_n, \state_{n+1}\right) \leftarrow  \mypar{1-\alpha_n} \, \underbrace{\reward\left(\state_n, \action_n, \state_{n+1}\right)}_{\text{old value}} \, + \, \alpha_n \, \cobj\left(\state_n, \action_n, \state_{n+1}\right),
\label{eq-rew_update}
\end{equation}
with $\alpha_n > 0$ the learning rate. For the TRs to be associated with the average contribution to the cost function of the transition-action $\mypar{\state_n\rightarrow \state_{n+1},\action_n}$, the learning rate is set to $\displaystyle \alpha_n = 1/N^{\mypar{i,l,j}}$, where $N^{\mypar{i,l,j}} \in \mathbb{N}$ corresponds to the number of times the transition-action $\left(\statei \rightarrow \state^{(j)}, \actionl\right)$ has occurred so far during the learning process.

\subsection{Reinforcement learning}
To derive a control strategy, one needs to determine a policy which would give the best control action given the current state of the system, as known through the hash function. The rewards associated with an action when in a given state have been learned and this information is now used to derive a control policy to drive the system along transitions and actions associated with the largest rewards.

When the probabilities of transition from a state to another are known, the policy may be identified by means of a dynamic programming algorithm,~\cite{Bellman1952,Mandl1974}. However, the distribution of transition probabilities and values \revision{$\left\{\expvalue_i\right\}_{i \in \indS}$} are difficult to reliably evaluate since neither the control policy nor the transition probabilities are stationary during the learning stage. In this situation, \emph{Reinforcement Learning} is a suitable class of methods for the control of Markov processes, \revision{\cite{Watkins1992,Powell2007,Lewis2009}.}

Among these methods, the \emph{Q-learning} approach consists in relying on the estimation of the \emph{Q-factors}, or \emph{action-values}, $Q^\pi$ which evaluate the expected reward of a state-action combination:
\revision{\begin{equation}
Q^\pi\mypar{i,l} := \left<\reward_i\mypar{\actionl}\right>+ \drate \, \sum_{j \in \Esph} p(i,\policy\mypar{i},j) \, \expvalue_j,
\label{eq-q_factor_def}
\end{equation}}
where \revision{$\left<\reward_i\mypar{\actionl}\right> := \Espoperator\left[\reward\left(\state_k = \statei, \actionl, \state_{k+1}\right)\right]$} is the empirical mean TR associated with applying the action \revision{$\actionl$} when in the state \revision{$\statei$. From Eq.~\eqref{eq-value_reform}, the action-value is hence the expected average reward of applying $\actionl$ while in state $\statei$ and subsequently following the policy $\policy$.}

As stated previously, the transition probabilities can not be accurately estimated. However, an iterative estimation of the Q-factors can be derived, \revision{\cite{Watkins1992,Lewis2009,Gosavi2011}.} Letting the initial Q-factors be given, the Q-factor associated with a state \revision{$\statei$ and an action $\actionl$} can be updated \revision{at time $t_n$} as follows:

\revision{\begin{eqnarray}
Q^\pi\mypar{i,l} & \leftarrow & \underbrace{Q^\pi\mypar{i,l}}_{\text{old value}} + \alpha_n \, \Delta Q^\pi, \\
\Delta Q^\pi & := & \mypar{ \reward\mypar{\state_n=\statei,\actionl,\state_{n+1}} + \drate \underbrace{\max_{\widetilde{l} \in \indA} Q^\pi\mypar{i,\widetilde{l}}}_{\text{``best'' value}} - \underbrace{Q^\pi\mypar{i,l}}_{\text{old value}} } \nonumber,
\label{eq-q_factor_estim}
\end{eqnarray}}
where $\alpha_n > 0$ is a learning factor. It can be shown that $Q^\pi$ converges to the true Q-factors when $n \rightarrow \infty$ if the following conditions hold, \cite{Watkins1992}: \textit{i)} the TRs $\reward$ are bounded, \textit{ii)} $0 < \alpha_n < 1$, $\forall n\in \mathbb{N}$, \textit{iii)} $\sum\limits_{n=1}^{\infty} \alpha_n \rightarrow \infty$ and \textit{iv)} $\sum\limits_{n=1}^{\infty} \mypar{\alpha_n}^2 < \infty$.

The action-value $Q^\pi\mypar{i,l}$ will increase when the reward associated with the 2-tuple $\left(\statei, \actionl\right)$ is good, \revision{\textit{i.e.}, such that $\Delta Q^\pi > 0$}, and decrease otherwise.
To learn a good policy, the system in different states is stimulated with different actions to estimate the Q-factors. The control policy is the action which, for each state, is associated with the largest Q-factor, \revision{\cite{Watkins1992,Lewis2009}.}

\revision{
\subsection{Robustness} \label{Sec_robustness}
A critical property of any realistic and practically useful control strategy is its resilience with respect to unpredictable events. These events encompass exogenous/external perturbations to the flow, sensor noise, actuator noise, etc. A control strategy robust to these perturbations is passive and brings the flow back to its nominal controlled state after the perturbation is gone.
As will be demonstrated in the application example below, see section~\ref{sssec-noise}, the present strategy is robust thanks to two properties:

First, the state of the system, as estimated via the LSH, is discrete. The locality sensitivity property of LSHs implies that a small perturbation $\gras{\epsilon}$ of the measurement vector $\gras{y}$ will likely result in the same key as the unperturbed measurement, $h^{\gras{v},\ql}\left(\gras{y} + \gras{\epsilon}\right) = h^{\gras{v},\ql}\left(\gras{y}\right)$. More precisely, the state estimation is strictly robust to any perturbation of energy $\left\|\gras{\epsilon}\right\|_2$ with probability
$\displaystyle \prod_{l=1}^{\Nv}{\max \left[0, 1 - \left\|\gras{\epsilon}\right\|_2 \, \lvert\cos\left(\angle\left(\gras{\epsilon}, \gras{v}_l\right)\right)\rvert \slash \ql \right]}$.

Second, the proposed method is also robust against perturbations of the flow dynamics. The learning of the control strategy relies on an ensemble average of rewards and Q-factors and then results in an optimal control policy in the ensemble mean-sense. Hence robustness against noise.}

\section{Results} \label{sec-results}

\subsection{Dynamical system}  \label{sec-results-Lorenz}
 To illustrate the methodology discussed above, we now consider the Lorenz 63 system~\cite{Lorenz1963}, defined by the following equations: 
\begin{equation}
\left\{
\begin{array}{lcl}
\frac{\text{d}X_1}{\text{d}t} & = & \sigma\mypar{X_2-X_1} + f_{X_1}\mypar{X,t}\\
\\
\frac{\text{d}X_2}{\text{d}t} & = & X_1\mypar{r-X_3} - X_2 + f_{X_2}\mypar{X,t}\\

\\\frac{\text{d}X_3}{\text{d}t} & = & X_1\times X_2 - b X_3 + f_{X_3}\mypar{X,t},\\
\end{array}\right.
\label{eq-lor_sys}
\end{equation}
with the common parameters $\mypar{\sigma,r,b} = \mypar{10,20,8/3}$. 
A chaotic attractor defines the dynamics, structured as two ``wings'' around two fixed points of the system. 
The state vector evolves on a wing, turning around a fixed point, before eventually jumping to the other wing. 
$f_{X_i}$ is the action on the component $i$ of $\bX := \left(X_1 \: X_2 \: X_3\right)$.
The chosen control objective is to remain on the ``left'' wing, $X_1 \le 0$.
\revision{The measure of the performance is given as the distance between the state vector and the left fixed point $\bX^\star$ of the system, $\drag := \left\|\bX - \bX^\star\right\|_2$, with $\bX^\star := \left(-\sqrt{b \left(r-1\right)}, -\sqrt{b \left(r-1\right)}, r-1\right)$.}
 
The observable $\obs$ is constructed from the time series $\mya{\obsdat\left(t - n \, \Delta t\right)}_{n = 0}^{\Ne-1}$ of $X_2$, sampled every $\Delta t = 0.025$ time units.
In this illustration, only $f_{X_1}$ is different from zero so that the Lorenz system is controlled only through the time-derivative $\text{d}X_1 \slash \text{d}t$ of the first component of its state vector. It mimics a realistic scenario, where sensors and actuators are distinct. Actuation values lie between $[-26,26]$ with a discretization step of $4$.
This leads to  $\Na = 14$ different actuations to control the system.
The cost function to minimize is given by Eq.~\eqref{eq-costfunc} \revision{and the control command is penalized with $\lagm = 0.1$.}

The embedding dimension is set to $\Ne = 8$. 
The first three singular vectors of a $\Ne \times \Nsnap$ Hankel matrix, built on $\Nsnap = 500$ measures $\mya{\obsdat\left(t - n \, \Delta t\right)}_{n = 0}^{\Nsnap-1}$, are used as the $\Nv = 3$ test vectors $\mya{\gras{v}_l}_{l=1}^{\Nv}$ of the LSHs. The quantization length is set to $\ql=45$ and leads to a cardinality of the set of keys of $\Nkey = 14$. 
\begin{figure*}[t]
\center
\subfigure[]{
\includegraphics[width=.55\linewidth]{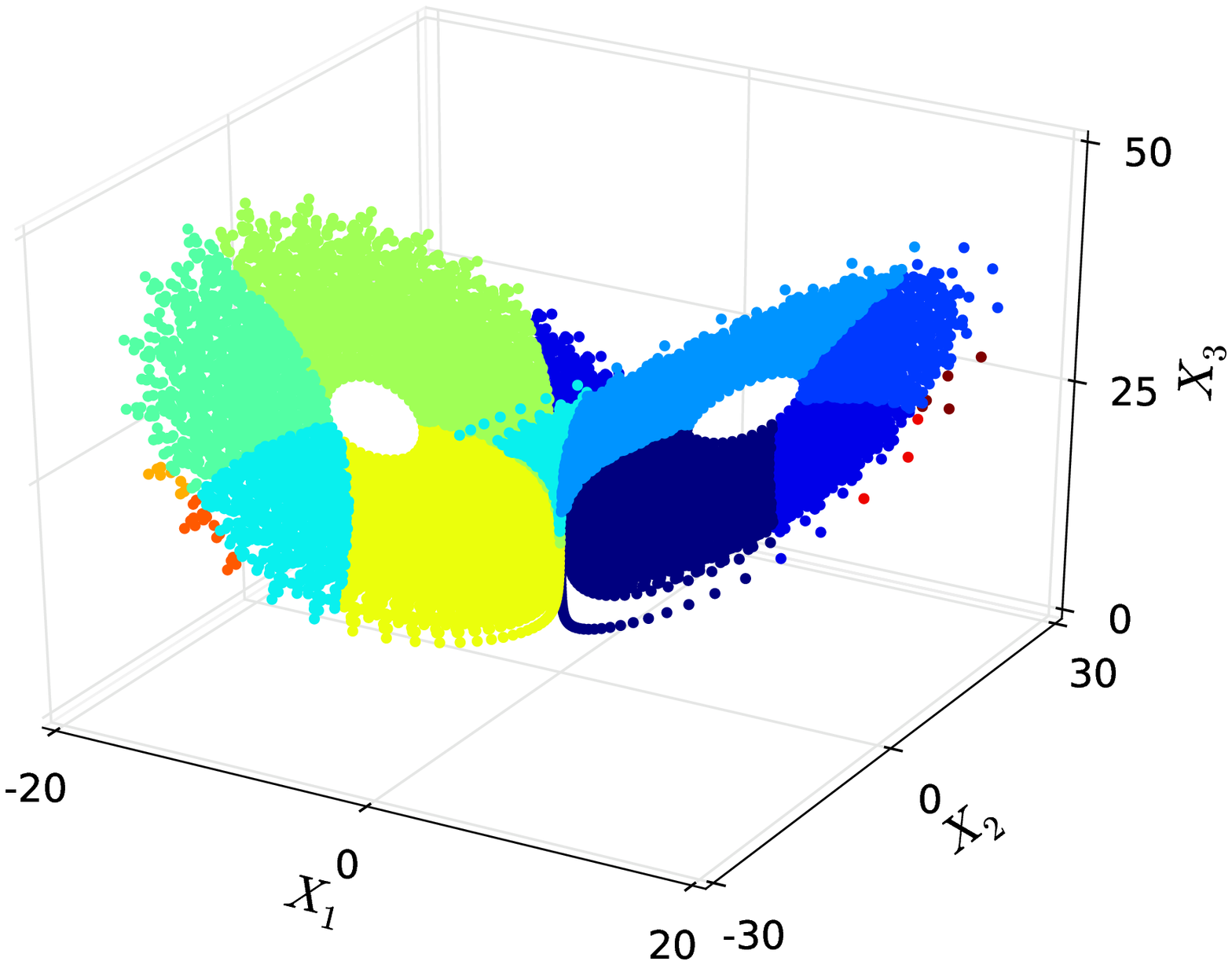}
\label{sfig-lor_clust}
}\\
\subfigure[]{
\includegraphics[width=.3\linewidth]{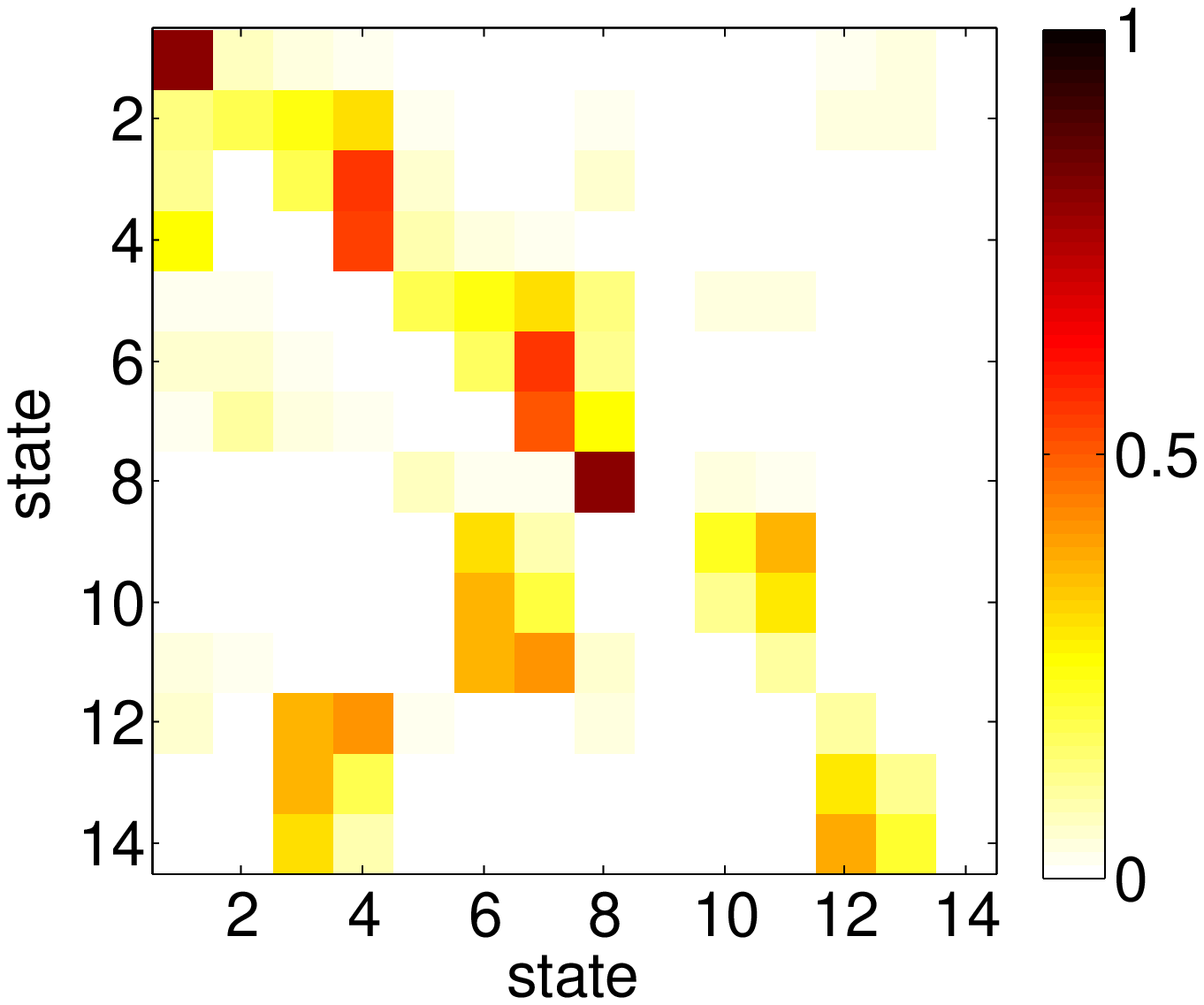}
\label{sfig-lor_trans}
}
\subfigure[]{
\includegraphics[width=.3\linewidth]{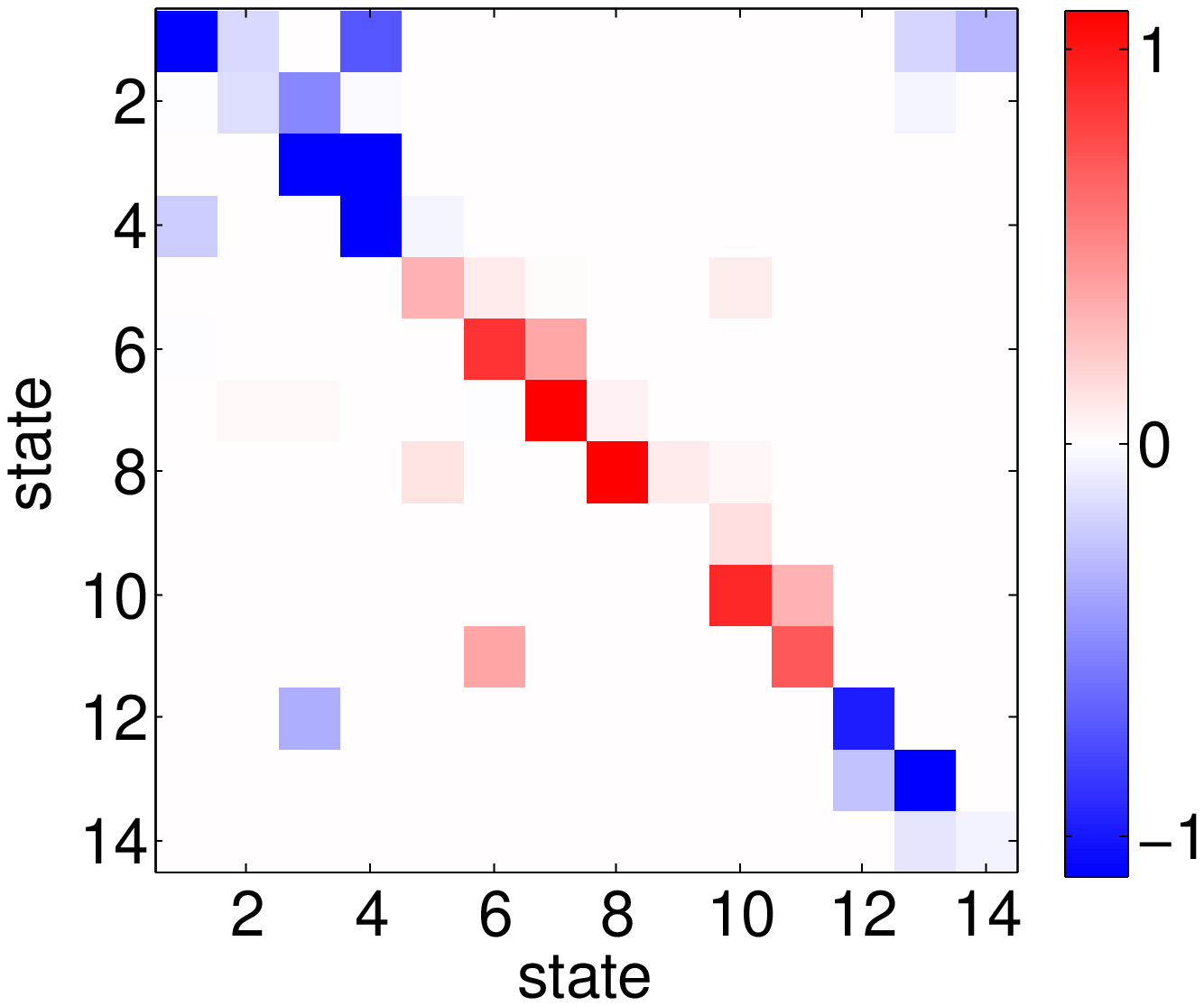}
\label{sfig-lor_rew}
}
\subfigure[]{
\includegraphics[width=.3\linewidth]{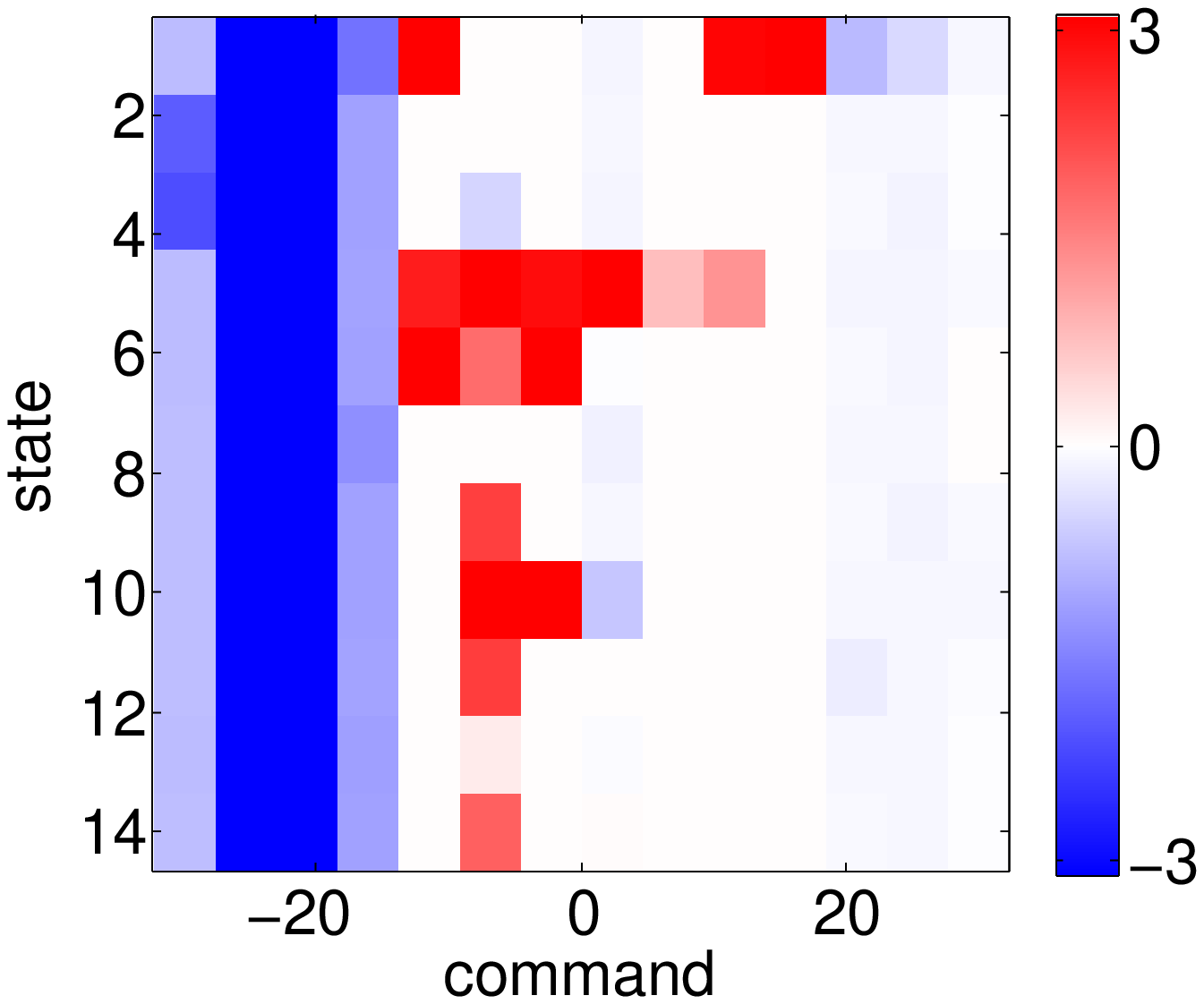}
\label{sfig-lor_Q}
}
\vspace{-1em}
\caption{\small \revision{(a): Identified clusters (color-coded) for the Lorenz 63 system.} (b): Mean state transition probabilities. The transition matrix has been iterated five times for readability. (c): Transition rewards associated with a state transition, for the \NULLtext{} command. The TRs have been rescaled between $-2$ and $2$ \revision{and the colormap saturated, in order to improve readability}. (d): Q-factors.}
\label{fig-lor_rewards}
\end{figure*}

\begin{figure*}[t]
\center
\subfigure[]{
\includegraphics[height=.3\linewidth]{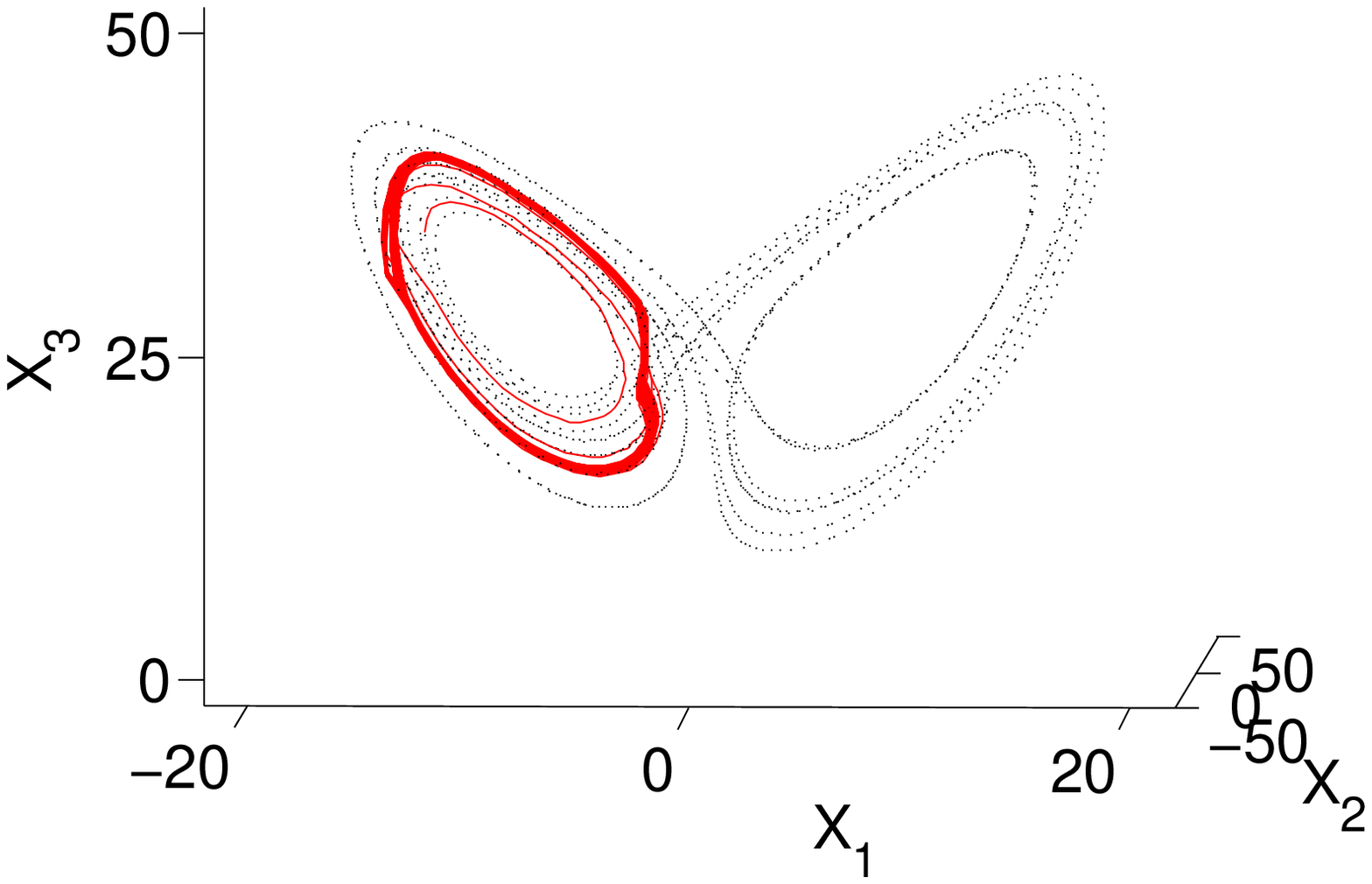}
\label{sfig-lor_att}
}
\hspace{3em}
\subfigure[]{
\includegraphics[height=.3\linewidth]{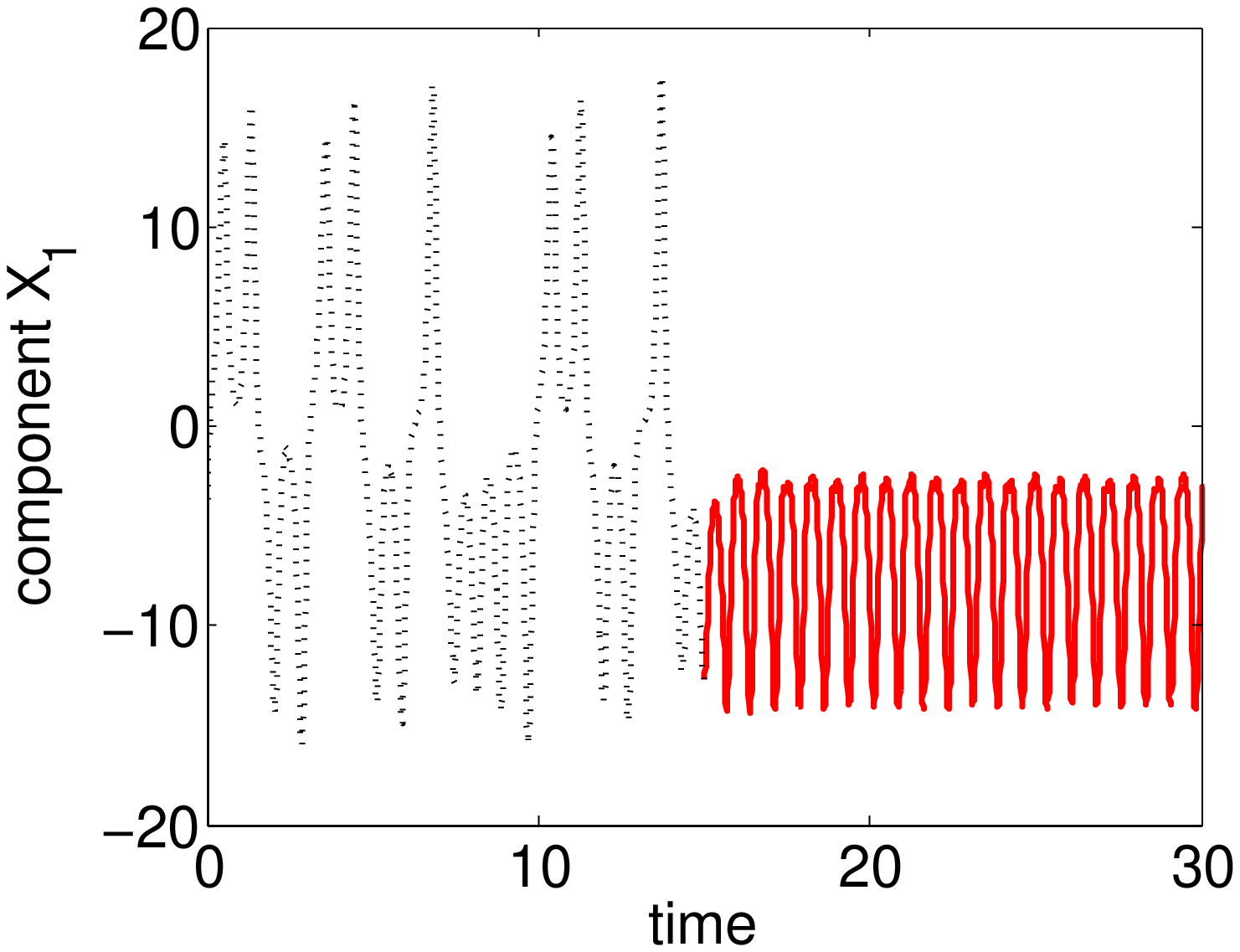}
\label{sfig-lor_stemp}
}
\vspace{-1em}
\caption{\small Uncontrolled Lorenz system in dashed black. Controlled response in solid red. The identified strategy is applied at time $t=15$. (a): State phase. (b): $X_1$ component.}
\label{fig-lor}
\end{figure*}

\begin{figure}
\begin{center}
\begin{tikzpicture}[scale=0.85] 


\node[draw,ellipse,fill=white] (k1) at (0,0) {1};
\node[draw,ellipse,fill=white] (k2) at (1,1) {2};
\node[draw,ellipse,fill=white] (k3) at (2,0) {3};
\node[draw,ellipse,fill=white] (k4) at (1,-1) {4};

\node[draw,ellipse,fill=white] (k6) at (5,0) {6};
\node[draw,ellipse,fill=white] (k7) at (6,1) {7};
\node[draw,ellipse,fill=white] (k8) at (7,0) {8};
\node[draw,ellipse,fill=white] (k5) at (6,-1) {5};

\node[draw,ellipse,fill=white] (k9) at (7,2) {\phantom{.}9\phantom{.}};
\node[draw,ellipse,fill=white] (k10) at (9,0) {10};
\node[draw,ellipse,fill=white] (k11) at (7,-2) {11};

\node[draw,ellipse,fill=white] (k12) at (0,2) {12};
\node[draw,ellipse,fill=white] (k13) at (-2,0) {13};
\node[draw,ellipse,fill=white] (k14) at (0,-2) {14};

\draw[->,>=latex,red!90] (k1) to[bend left] (k2);
\draw[->,>=latex,red!30] (k1) to[bend left] (k13);

\draw[->,>=latex,red!90] (k2) to[bend left] (k4);
\draw[->,>=latex,red!30] (k2) to[bend left] (k3);

\draw[->,>=latex,red!90] (k3) to[bend left] (k4);
\draw[->,>=latex,red!30] (k3) to[bend left] (k1);

\draw[->,>=latex,red!90] (k4) to[bend left] (k1);
\draw[->,>=latex,red!30] (k4) to[bend right] (k5);

\draw[->,>=latex,red!90] (k5) to[bend left] (k7);
\draw[->,>=latex,red!30] (k5) to[bend left] (k6);

\draw[->,>=latex,red!90] (k6) to[bend left] (k7);
\draw[->,>=latex,red!30] (k6) to[bend left] (k8);

\draw[->,>=latex,red!90] (k7) to[bend left] (k8);
\draw[->,>=latex,red!30] (k7) to[bend right] (k2);

\draw[->,>=latex,red!90] (k8) to[bend left] (k5);
\draw[->,>=latex,red!30] (k8) to[bend right] (k10);

\draw[->,>=latex,red!90] (k9) to[bend left] (k11);
\draw[->,>=latex,red!30] (k9) to[bend right] (k6);

\draw[->,>=latex,red!90] (k10) to[bend right] (k6);
\draw[->,>=latex,red!30] (k10) to[bend left] (k11);

\draw[->,>=latex,red!90] (k11) to[bend left] (k7);
\draw[->,>=latex,red!30] (k11) to[bend left] (k6);

\draw[->,>=latex,red!90] (k12) to[bend left] (k4);
\draw[->,>=latex,red!30] (k12) to[bend left] (k3);

\draw[->,>=latex,red!90] (k13) to[bend left] (k3);
\draw[->,>=latex,red!30] (k13) to[bend left] (k12);

\draw[->,>=latex,red!90] (k14) to[bend left] (k12);
\draw[->,>=latex,red!30] (k14) to[bend right] (k3);

\node[draw,ellipse,fill=white] (k1) at (0,0) {1};
\node[draw,ellipse,fill=white] (k2) at (1,1) {2};
\node[draw,ellipse,fill=white] (k3) at (2,0) {3};
\node[draw,ellipse,fill=white] (k4) at (1,-1) {4};

\node[draw,ellipse,fill=white] (k6) at (5,0) {6};
\node[draw,ellipse,fill=white] (k7) at (6,1) {7};
\node[draw,ellipse,fill=white] (k8) at (7,0) {8};
\node[draw,ellipse,fill=white] (k5) at (6,-1) {5};

\node[draw,ellipse,fill=white] (k9) at (7,2) {\phantom{.}9\phantom{.}};
\node[draw,ellipse,fill=white] (k10) at (9,0) {10};
\node[draw,ellipse,fill=white] (k11) at (7,-2) {11};

\node[draw,ellipse,fill=white] (k12) at (0,2) {12};
\node[draw,ellipse,fill=white] (k13) at (-2,0) {13};
\node[draw,ellipse,fill=white] (k14) at (0,-2) {14};

\draw[->,dashed,>=latex,red!90] (k12) to[bend left] (k4);
\draw[->,dashed,>=latex,red!90] (k11) to[bend left] (k7);

\end{tikzpicture}
\end{center}
\vspace{-1em}
\caption{\small \revision{Transitions between clusters for the Lorenz system. Only the first two most probable transitions are represented. The dark red arrows represent the most probable transitions from one cluster to another, while the pale red represents the second most probable transitions.}}
\label{fig-schem_lor}
\end{figure}
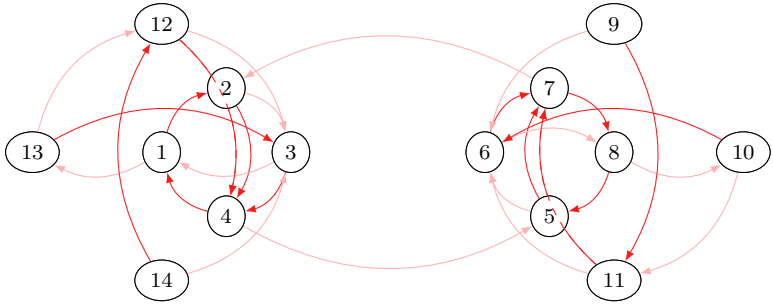

Once the test vectors are determined from the Hankel matrix, the states are to be identified via the LSHs, see Fig.~\ref{sfig-lor_clust}.
It is hence possible to infer both the dynamics of the system from the state transitions and the rewards associated with the objective.
The mean transition probabilities from one state to another, after five time increments in order to improve visualization, are plotted in Fig.~\ref{sfig-lor_trans}.
As expected for the Lorenz system, the dynamics are seen to be driven by two main cycles, \revision{see Fig.~\ref{fig-schem_lor}.}
There are two statistically dominant sequences of transitions cycling around all states between 1 (resp. 5) and 4 (resp. 8), corresponding to the right (resp. left) wing of the Lorenz attractor. States 9 to 11 and 12 to 14 represent sub-transitions from a main cluster to another one.

From the observations of the system, one learns the $\Nkey \times \Na \times \Nkey \simeq 2100$ transition rewards matrix $\reward$ and the $\Nkey \times \Na \simeq 150$ Q-factors matrix $Q$ by repeatedly applying Eq.~\eqref{eq-rew_update} and Eq.~\eqref{eq-q_factor_estim}.
The transition matrix is sparse, see Fig.~\ref{sfig-lor_trans}, and many transitions hence hardly ever occur. 
Thus, the learned TRs and the Q-factors matrices are also sparse, see Fig.~\ref{sfig-lor_rew} and Fig.~\ref{sfig-lor_Q}. 
The difference between one wing and the other is clearly discernible in the rewards, the first block being associated with negative rewards while the second block is associated with positive rewards.
By construction of the rewards, the use of a strong (expensive) command is also discouraged, as can be seen in the Q-factors, Fig.~\ref{sfig-lor_Q}.
The identified control strategy succeeds in staying on the left wing, see Fig.~\ref{fig-lor}.

\subsection{Two-dimensional numerical flow}
To further illustrate the methodology discussed above, consider a 2-D laminar flow around a circular cylinder, in two situations: a fixed, or random in time angle of the incoming flow.

\subsubsection{Configuration of the test case} \label{sssec-config}
The considered Reynolds number of the flow is $Re = 200$ based on the cylinder diameter and the upstream flow velocity. Details of the simulation can be found in \cite{Lemaitre2003}. The observable $\obs$ is constructed from the time series $\mya{\obsdat\left(t - n \, \Delta t\right)}_{n = 0}^{\Ne-1}$ of a single pressure sensor, sampled every $\Delta t = 0.75$ time units. This sensor is located on the cylinder surface at an angle of $160$ degrees from the upstream stagnation point when the angle is fixed. In this case, the pressure signal oscillates with a period of about 9 time units and half as much for the drag.

Actuation, \textit{i.e.}, control of the flow is achieved via blowing or suction through the whole cylinder surface. 
Actuation values lie within $[-0.2,0.02]$, with a discretization step of $0.02$, ranging from suction to blowing. It leads to $\Na = 12$ different actuations to control the system.
The cost function to minimize is given in Eq.~\eqref{eq-costfunc}, with $\lagm = 23$. \revision{It corresponds to the drag ($\drag$ is the drag coefficient) induced by the cylinder penalized with the intensity of the command. The penalty $\lagm$ was chosen so that the resulting command remains within the operating range of the actuator.}
The embedding dimension is set to $\Ne = 14$. 
The first five singular vectors of a $\Ne \times \Nsnap$ Hankel matrix, built on $\Nsnap = 500$ measures $\mya{\obsdat\left(t - n \, \Delta t\right)}_{n = 0}^{\Nsnap-1}$, are used as the $\Nv = 5$ test vectors $\mya{\gras{v}_l}_{l=1}^{\Nv}$ of the LSHs. The quantization length is set to $\ql=50$ and leads to a cardinality of the set of keys of $\Nkey = 15$. 

\begin{figure*}[t]
\center
\subfigure[]{
\includegraphics[height=.3\linewidth]{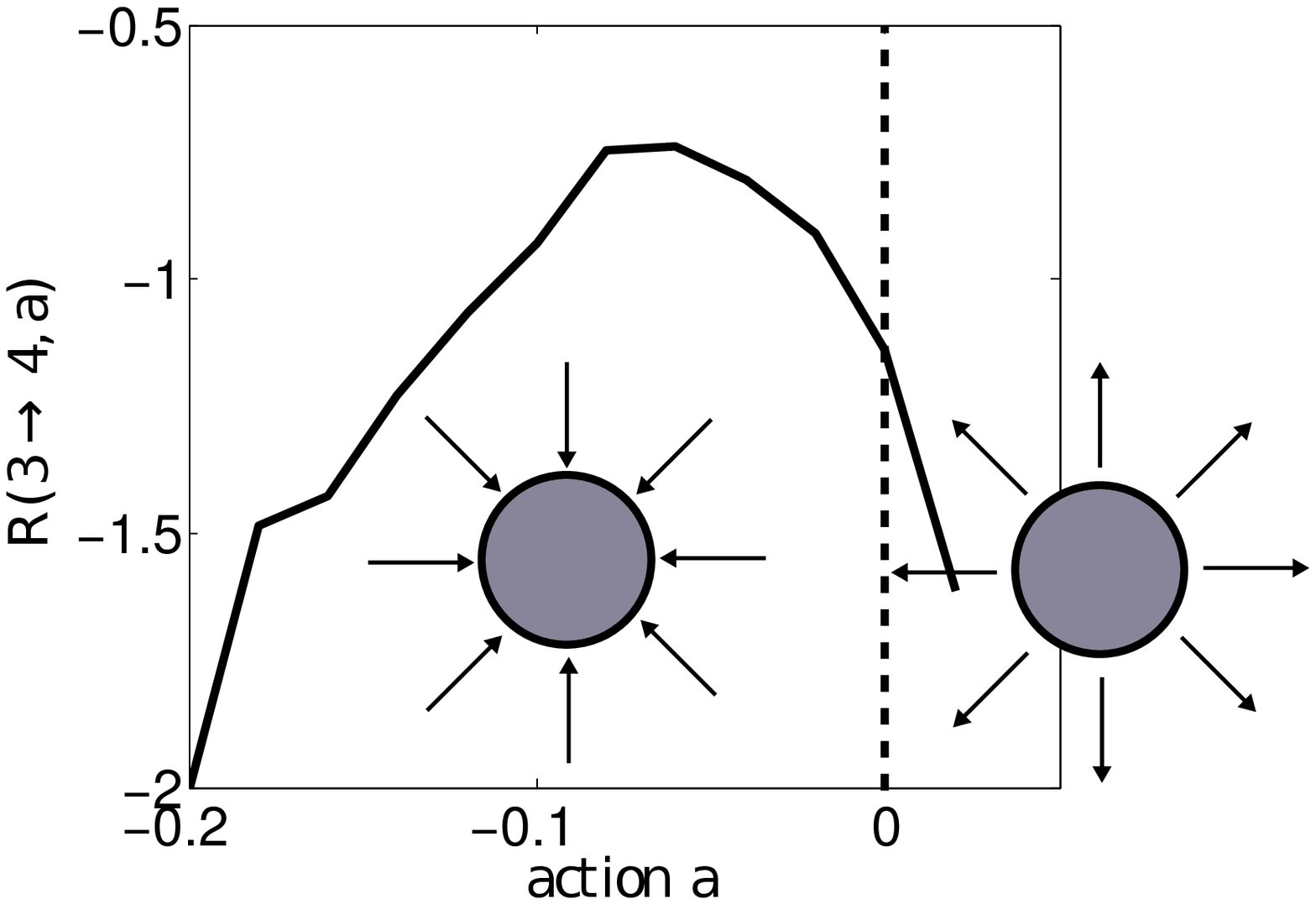}
\label{sfig-ra}
}
\hspace{3em}
\subfigure[]{
\includegraphics[height=.3\linewidth]{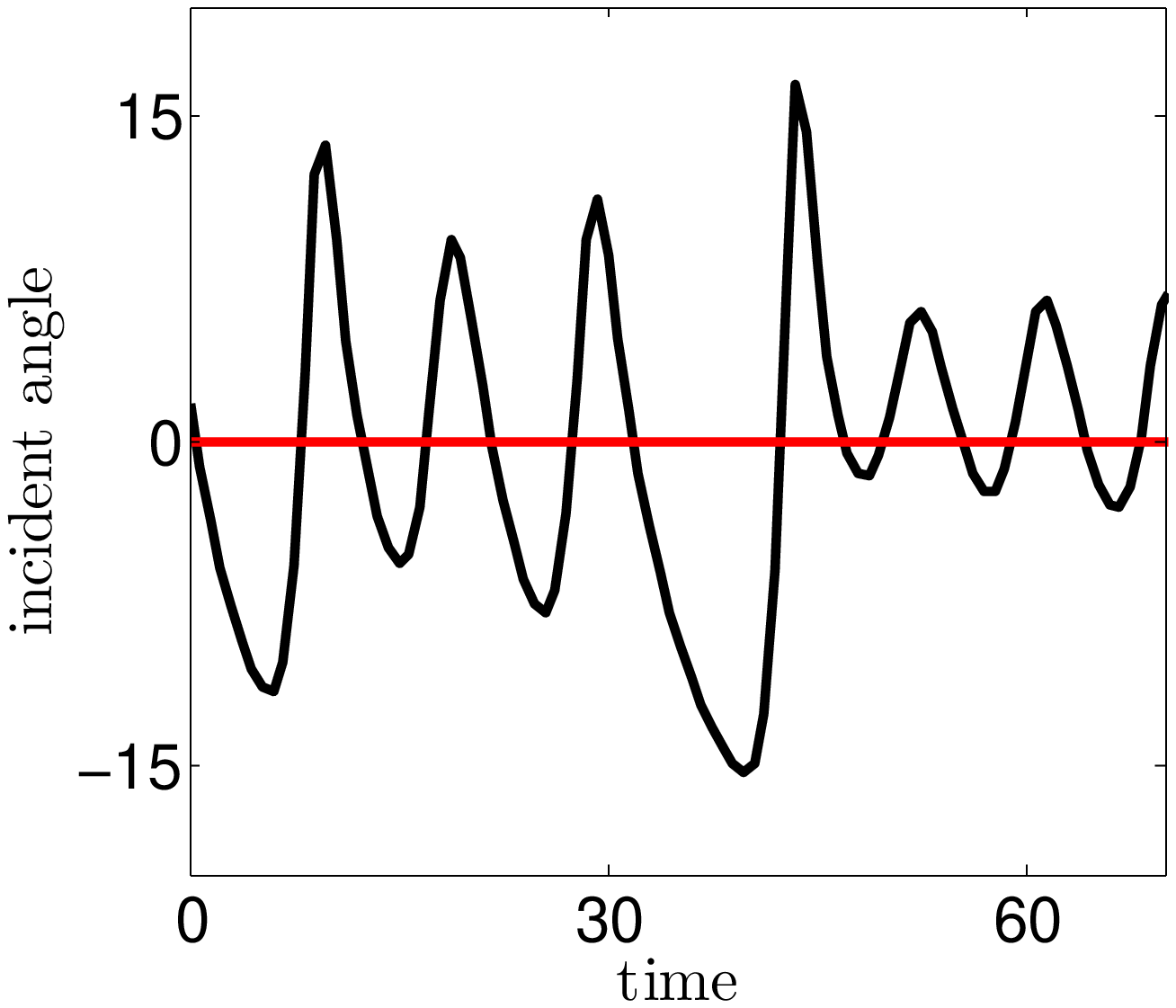}
\label{sfig-angle}
}
\vspace{-1em}
\caption{\small (a): Transition rewards associated to the transition $3\rightarrow 4$, with respect to the command. The TRs have been rescaled between $-2$ and $2$. The drawings illustrate the actuation (suction or blowing). (b): Time-evolution of the incident angle of the incoming flow, for the noiseless (red) and noisy (black) case.}
\label{fig-div}
\end{figure*}

Two situations are considered to illustrate the proposed control algorithm. First, the incident angle of the incoming flow is kept constant at its zero nominal value, Sec.~\ref{sssec-nonoise}. Alternatively, in Sec.~\ref{sssec-noise}, the angle is a random process, smooth in time, whose realizations range from -20 to 20 degrees around the nominal value with a uniform probability distribution, see Fig.~\ref{sfig-angle}.

\subsubsection{Noiseless case: nominal incidence} \label{sssec-nonoise}
\begin{figure}[t]
\center
\subfigure[]{
\includegraphics[width=.3\linewidth]{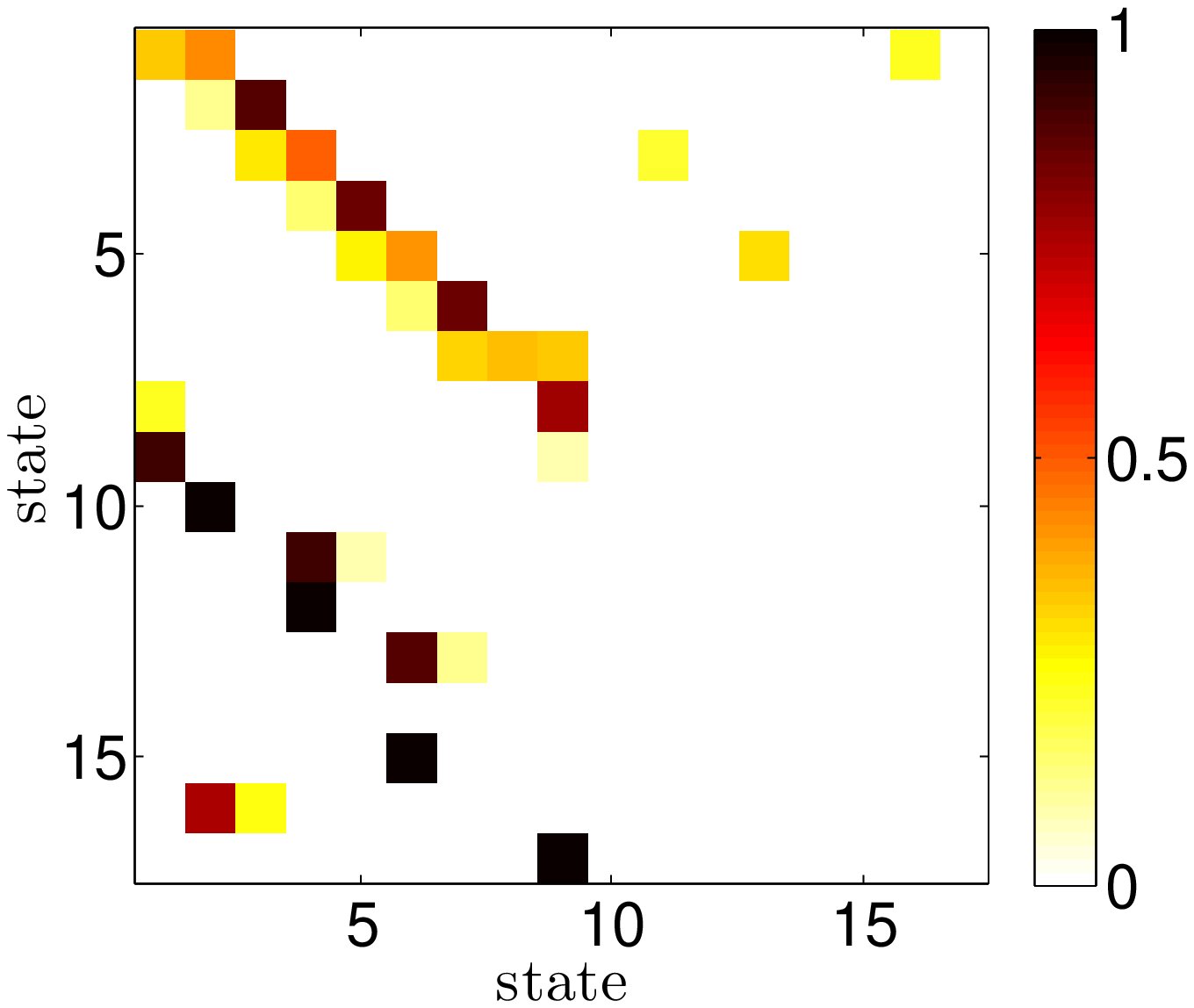}
\label{sfig-trans}
}
\subfigure[]{
\includegraphics[width=.3\linewidth]{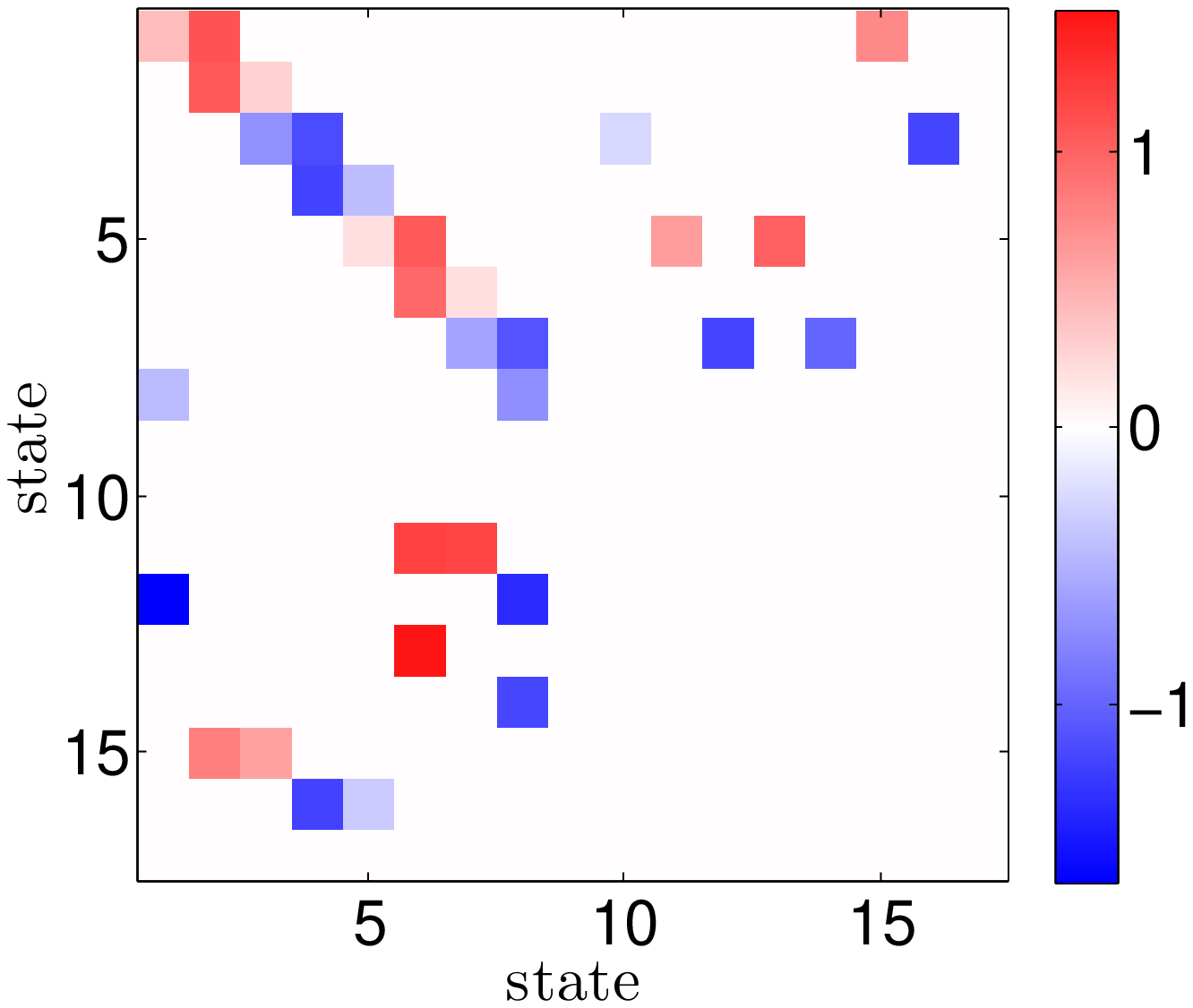}
\label{sfig-rew}
}
\subfigure[]{
\includegraphics[width=.3\linewidth]{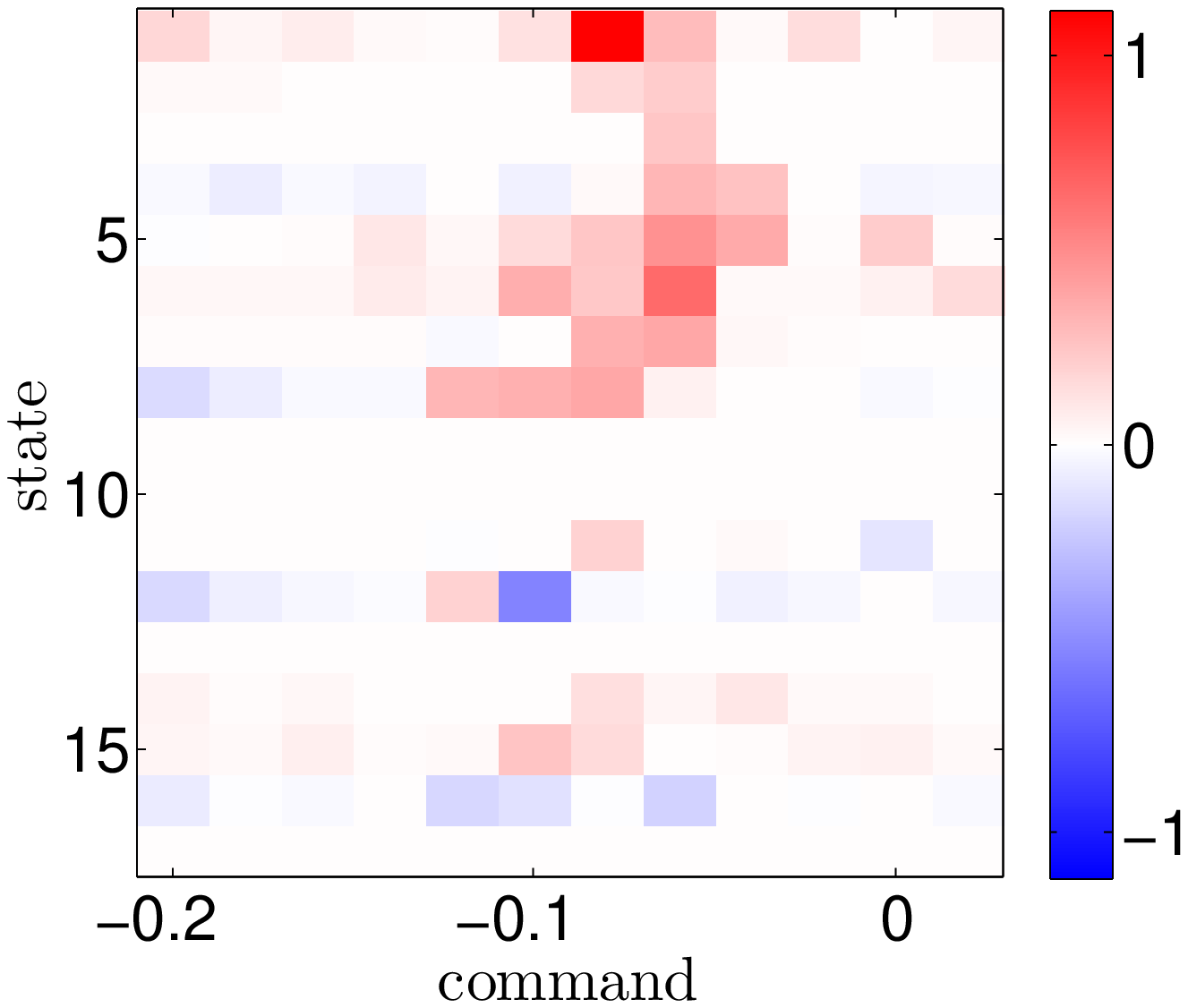}
\label{sfig-Q}
}
\vspace{-1em}
\caption{\small (a): Mean state transition probabilities. (b): Transition rewards associated with a state transition, for the \NULLtext{} command. The TRs have been rescaled between $-2$ and $2$ \revision{and the colormap saturated, in order to improve readability}. (c): Q-factors.}
\label{fig-rewards}
\end{figure}

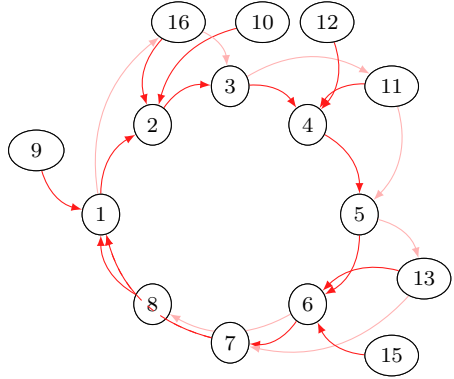
\begin{figure}
\begin{center}
\begin{tikzpicture}[scale=0.85] 


\node[draw,ellipse,fill=white] (k1) at (0,0) {1};
\node[draw,ellipse,fill=white] (k2) at (.8,1.4) {2};
\node[draw,ellipse,fill=white] (k3) at (2,2) {3};
\node[draw,ellipse,fill=white] (k4) at (3.2,1.4) {4};
\node[draw,ellipse,fill=white] (k5) at (4,0) {5};
\node[draw,ellipse,fill=white] (k6) at (3.2,-1.4) {6};
\node[draw,ellipse,fill=white] (k7) at (2,-2) {7};
\node[draw,ellipse,fill=white] (k8) at (.8,-1.4) {8};

\node[draw,ellipse,fill=white] (k9) at (-1,1) {\phantom{.}9\phantom{.}};
\node[draw,ellipse,fill=white] (k10) at (2.5,3) {10};
\node[draw,ellipse,fill=white] (k11) at (4.5,2) {11};
\node[draw,ellipse,fill=white] (k12) at (3.5,3) {12};
\node[draw,ellipse,fill=white] (k13) at (5,-1) {13};
\node[draw,ellipse,fill=white] (k15) at (4.5,-2.2) {15};
\node[draw,ellipse,fill=white] (k16) at (1.2,3) {16};

\draw[->,>=latex,red!90] (k1) to[bend left] (k2);
\draw[->,>=latex,red!30] (k1) to[bend left] (k16);

\draw[->,>=latex,red!90] (k2) to[bend left] (k3);

\draw[->,>=latex,red!90] (k3) to[bend left] (k4);
\draw[->,>=latex,red!30] (k3) to[bend left] (k11);

\draw[->,>=latex,red!90] (k4) to[bend left] (k5);

\draw[->,>=latex,red!90] (k5) to[bend left] (k6);
\draw[->,>=latex,red!30] (k5) to[bend left] (k13);

\draw[->,>=latex,red!90] (k6) to[bend left] (k7);
\draw[->,>=latex,red!30] (k6) to[bend left] (k8);

\draw[->,>=latex,red!90] (k7) to[bend left] (k1);

\draw[->,>=latex,red!90] (k8) to[bend left] (k1);

\draw[->,>=latex,red!90] (k9) to[bend right] (k1);

\draw[->,>=latex,red!90] (k10) to[bend right] (k2);

\draw[->,>=latex,red!90] (k11) to[bend right] (k4);
\draw[->,>=latex,red!30] (k11) to[bend left] (k5);

\draw[->,>=latex,red!90] (k12) to[bend left] (k4);

\draw[->,>=latex,red!90] (k13) to[bend right] (k6);
\draw[->,>=latex,red!30] (k13) to[bend left] (k7);

\draw[->,>=latex,red!90] (k15) to[bend left] (k6);

\draw[->,>=latex,red!90] (k16) to[bend right] (k2);
\draw[->,>=latex,red!30] (k16) to[bend left] (k3);

\node[draw,ellipse,fill=white] (k1) at (0,0) {1};
\node[draw,ellipse,fill=white] (k2) at (.8,1.4) {2};
\node[draw,ellipse,fill=white] (k3) at (2,2) {3};
\node[draw,ellipse,fill=white] (k4) at (3.2,1.4) {4};
\node[draw,ellipse,fill=white] (k5) at (4,0) {5};
\node[draw,ellipse,fill=white] (k6) at (3.2,-1.4) {6};
\node[draw,ellipse,fill=white] (k7) at (2,-2) {7};
\node[draw,ellipse,fill=white] (k8) at (.8,-1.4) {8};

\node[draw,ellipse,fill=white] (k9) at (-1,1) {\phantom{.}9\phantom{.}};
\node[draw,ellipse,fill=white] (k10) at (2.5,3) {10};
\node[draw,ellipse,fill=white] (k11) at (4.5,2) {11};
\node[draw,ellipse,fill=white] (k12) at (3.5,3) {12};
\node[draw,ellipse,fill=white] (k13) at (5,-1) {13};
\node[draw,ellipse,fill=white] (k15) at (4.5,-2.2) {15};
\node[draw,ellipse,fill=white] (k16) at (1.2,3) {16};

\draw[->,dashed,>=latex,red!90] (k7) to[bend left] (k1);

\end{tikzpicture}
\end{center}
\vspace{-1em}
\caption{\small \revision{Transitions between clusters for the cylinder flow. Only the first two most probable transitions are represented. The dark red arrows represent the most probable transitions from a cluster to another one, while the pale red represents the second most probable transitions.}}
\label{fig-schem_cyl}
\end{figure}

Once the test vectors are determined from the Hankel matrix, and the states are identified via $\mathfrak{h}$, it is possible to infer the dynamics of the system and the associated rewards. The learned transition reward $\reward\mypar{3,\actionl,4}$ is plotted in Fig.~\ref{sfig-ra}. It clearly exhibits a maximum which corresponds to the best compromise between a sufficient decrease of $\drag$ while a reasonable increase of $\monabs{\action}^2$. The estimated transition probabilities from one state to another are plotted in Fig.~\ref{sfig-trans}.
In the present case, the dynamics are seen to be rather periodic, with a statistically dominant sequence of transitions cycling around all states between 1 and 8\revision{, see Fig.~\ref{fig-schem_cyl}}.
Other states are ``transient'' states between two stages of the main cycle. For instance, a transition from state $1$ to state $16$ can occur with a low probability but the next transition will be to state $2$ (or, less likely, state $3$).
Further analysis of the transition matrix can give more insights into the dynamics and on the relevance of other sequences, \emph{e.g.}, via stability analysis,~\cite{Kaiser2014}, symbolic dynamics based on keys, \cite{Lusseyran2008}, or the Kullback-Leibler entropy,~\cite{Kaiser2014}.

From the observations of the system, one learns the $\Nkey \times \Na \times \Nkey \simeq 2700$ transition rewards matrix $\reward$ and the $\Nkey \times \Na \simeq 180$ Q-factors matrix $Q$ by repeatedly applying Eq.~\eqref{eq-rew_update} and Eq.~\eqref{eq-q_factor_estim}.
As in the Lorenz system case (see above in Sec.~\ref{sec-results-Lorenz}), the transition matrix is sparse, as can be appreciated from Fig.~\ref{sfig-trans}, and many transitions hence hardly ever occur. 
Thus, the learned TRs and the Q-factors matrices are also sparse, see Fig.~\ref{sfig-rew} and Fig.~\ref{sfig-Q}.

The performance of the control strategy is assessed in terms of the performance indicator $\effic$ defined as the difference between the time-averaged cost $\left<\costfunction\right> := \left<\costfunction\left(t\right)\right>_t$ and the time-averaged cost with a \NULLtext{} strategy ($a = 0$), $\left<\costfunction_{\NULL}\right>$. The performance indicator is scaled with the difference between $\left<\costfunction_{\NULL}\right>$ and the cost $\left<\costfunction_{\oracle}\right>$ of \revision{an optimal time-invariant control strategy}:
\begin{equation}
\effic := \mypar{\left<\costfunction_{\NULL}\right> - \left<\costfunction\right>} / \mypar{\left<\costfunction_{\NULL}\right> - \left<\costfunction_{\oracle}\right>}.
\label{eq-effic}
\end{equation}

{
The evolution of $\effic$ as a function of the learning effort, \revision{defined as the number $\Nsnap$ of measurements during the learning phase, is plotted in Fig.~\ref{sfig-dist}.}
The performance indicator increases with the amount of learning for computing the TRs, but quickly reaches a plateau. 
The TR matrix being sparse, the information of the learning stage focuses onto a limited number of unknowns and the Q-factors quickly converge. 

As expected, the values \revision{$\left\{\expvalue_i\right\}_{i \in \indS}$}, computed with the learned policies are larger than the values computed with the \oracletext{} policy, on average, see Tab.~\ref{tab-nonoise_values}.
\begin{table}[h]
\begin{center}
\revision{
\begin{tabular}{l||c|c|c}
Policy & \NULLtext{} & \oracletext{} & Present strategy \\
\hline
Average value $\left<\expvalue\right>$ & -15.8 &  9.5 & 13.2
\end{tabular}
\caption{Average expected value $\left<\expvalue\right> := \Espoperator_{i \in \indS}\left[\expvalue_i\right]$ associated with the \NULLtext{}, \oracletext{} and present strategy policies. The learning effort for the rewards (resp. the Q-factors) was $10000$.}
\label{tab-nonoise_values}
}
\end{center}
\end{table}

\begin{figure}[t]
\center
\subfigure[]{
\includegraphics[height=.23\linewidth]{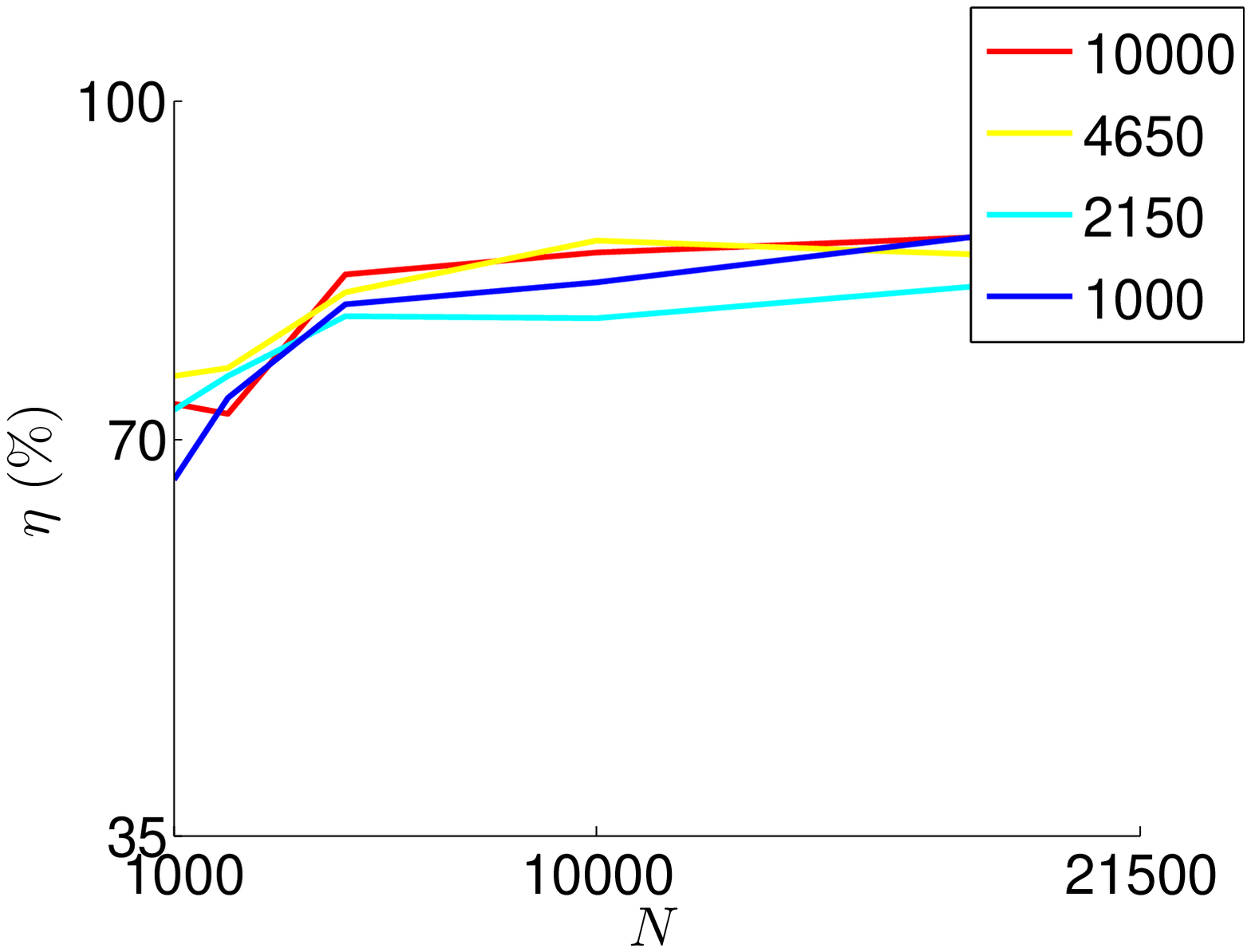}
\label{sfig-dist}
}
\subfigure[]{
\includegraphics[height=.23\linewidth]{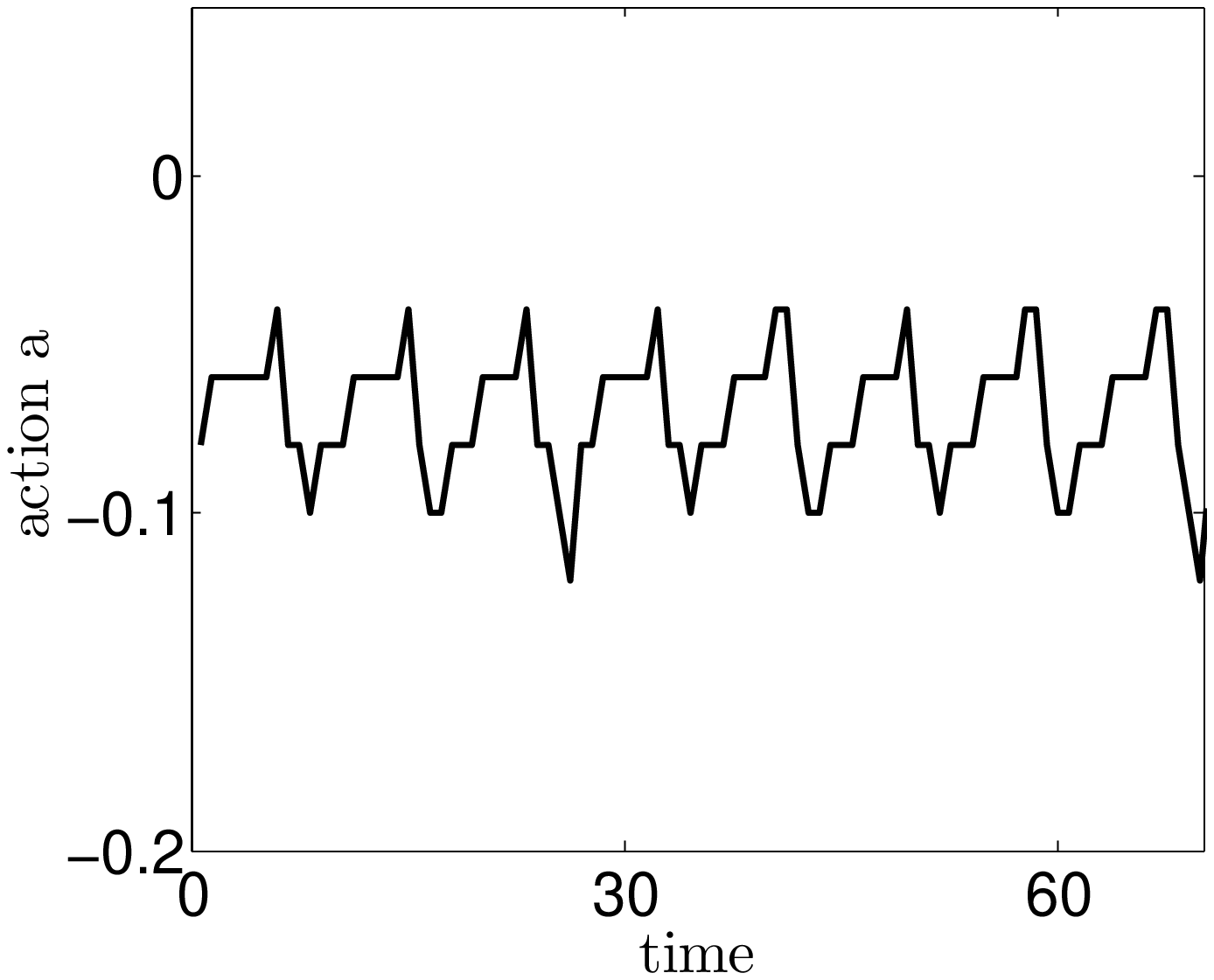}
\label{sfig-com}
}
\subfigure[]{
\includegraphics[height=.23\linewidth,width=.30\linewidth]{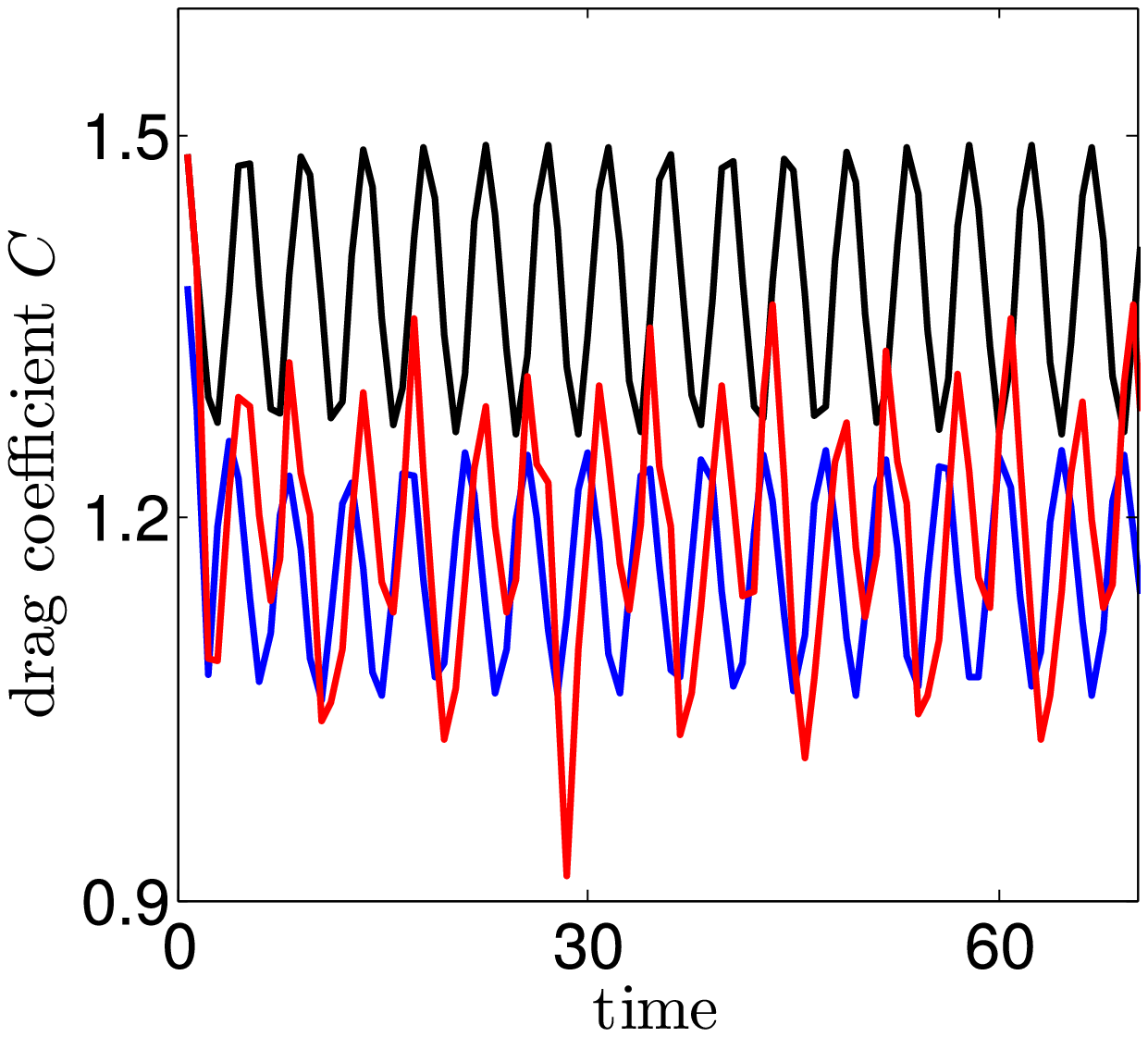}
\label{sfig-drag}
}
\vspace{-1em}
\caption{\small Illustration of the control policy. (a): Performance indicator with respect to the learning effort of the transition rewards matrix $\reward$. Colors (from blue to red) encode different efforts in learning $Q$. (b): Actuation as a function of time, $\action(t)$. (c): Time-evolution of the drag coefficient under three different control policies: zero command (\NULLtext{}, black), \oracletext{} strategy (blue), and the \revision{optimal policy from the} present approach (red).}
\label{fig-drag}
\end{figure}

The resulting control command is plotted in Fig.~\ref{sfig-com} and is seen to exhibit an oscillatory behavior. The associated drag coefficient of the cylinder flow is plotted in Fig.~\ref{sfig-drag} for the present approach as well the \NULLtext{} and the \oracletext{} strategy. \revision{The identified control is seen to perform well. The drag coefficient for the identified control resembles the one given by the \oracletext{} control.}

To illustrate the impact of the control on the system, the time-averaged pressure field around the cylinder is plotted in Fig.~\ref{fig-pressure}, both with and without control. When control is applied, the pressure difference between the upstream stagnation point and the rear cylinder vicinity is significantly reduced. Further insights about the control effect can be gained by examining Fig.~\ref{fig-streamlines} where the time-averaged streamlines are plotted. With control, \textit{i.e.}, with a negative normal velocity at the cylinder surface, small recirculation bubbles significantly weaken, or even vanish, and the separation of the boundary layer from the cylinder surface is postponed further downstream, reducing the effective width of the wake.
The length $L_r$ of the recirculation bubble drops from $L_r = 1.11$ in the case of the \NULLtext{} command (no control), see Fig.~\ref{sfig-pnocont}, to $L_r = 0.95$ when the command identified by the present control approach is applied, see Fig.~\ref{sfig-pcont}. The length of the recirculation bubble is hence reduced by $15\%$. Suction at the cylinder surface tends to slightly increase the viscous drag (thinner boundary layers with larger velocity gradients) but significantly decreases the width of the wake and the pressure defect at the back of the cylinder, producing a lower drag.
\begin{figure}[t]
\center
\subfigure[]{
\includegraphics[width=.65\linewidth]{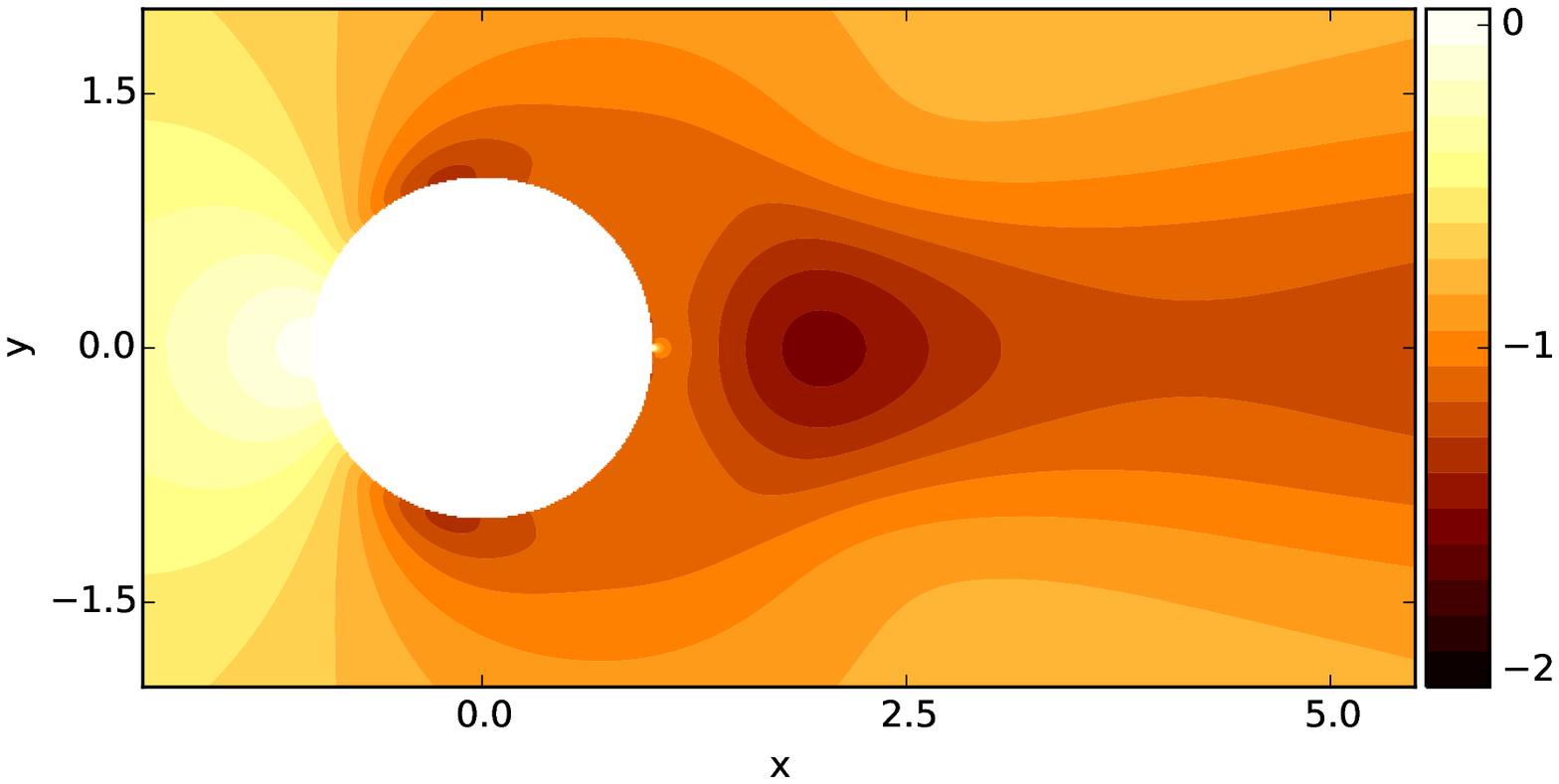}
\label{sfig-pnocont}
}\\
\subfigure[]{
\includegraphics[width=.65\linewidth]{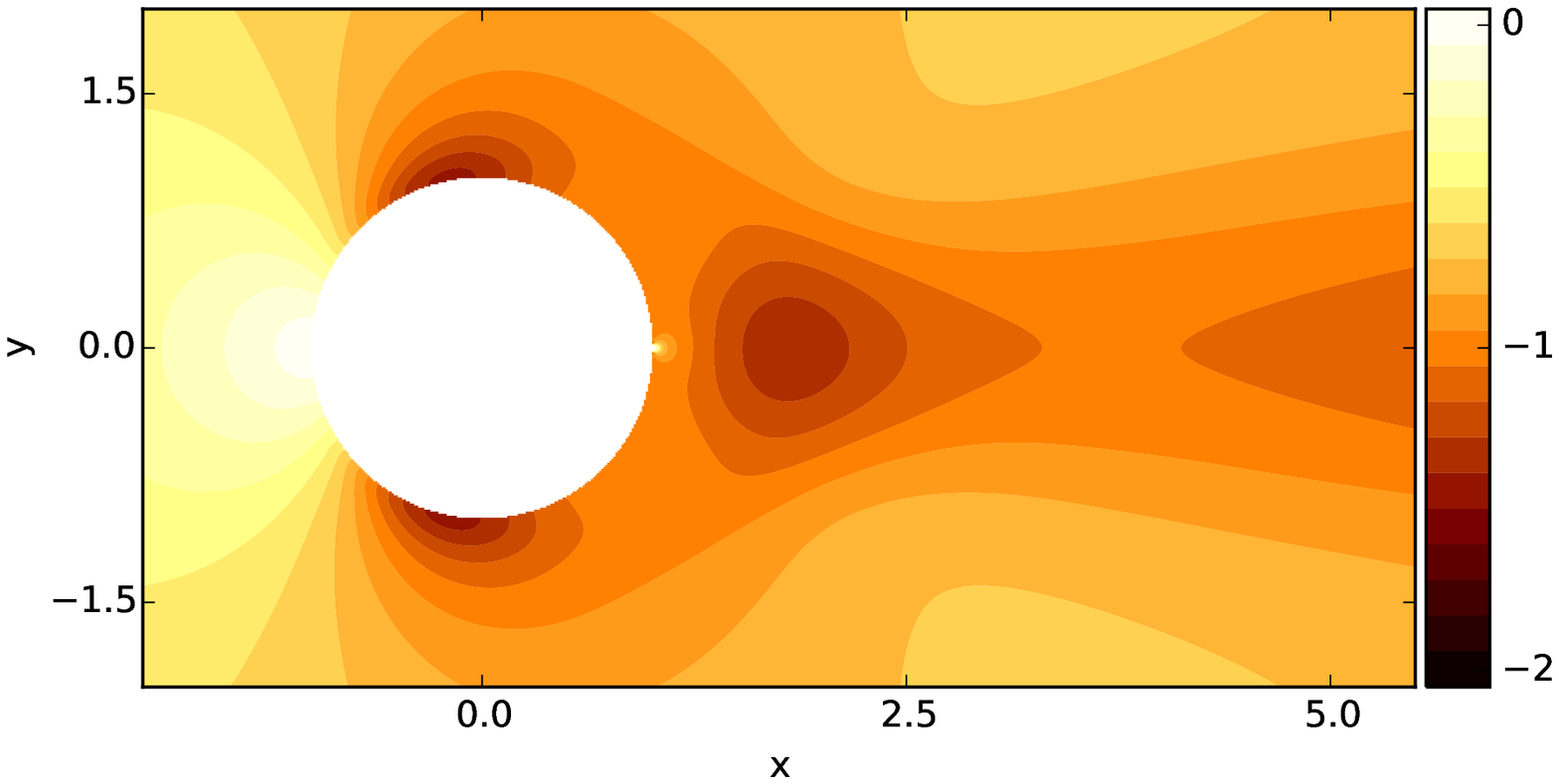}
\label{sfig-pcont}
}
\vspace{-1em}
\caption{\small Mean pressure field. (a): for the \NULLtext{} command. (b): for the present identified control strategy. Note that only a part of the computational domain is plotted.}
\label{fig-pressure}
\end{figure}

\begin{figure}[t]
\center
\subfigure[]{
\includegraphics[width=.65\linewidth]{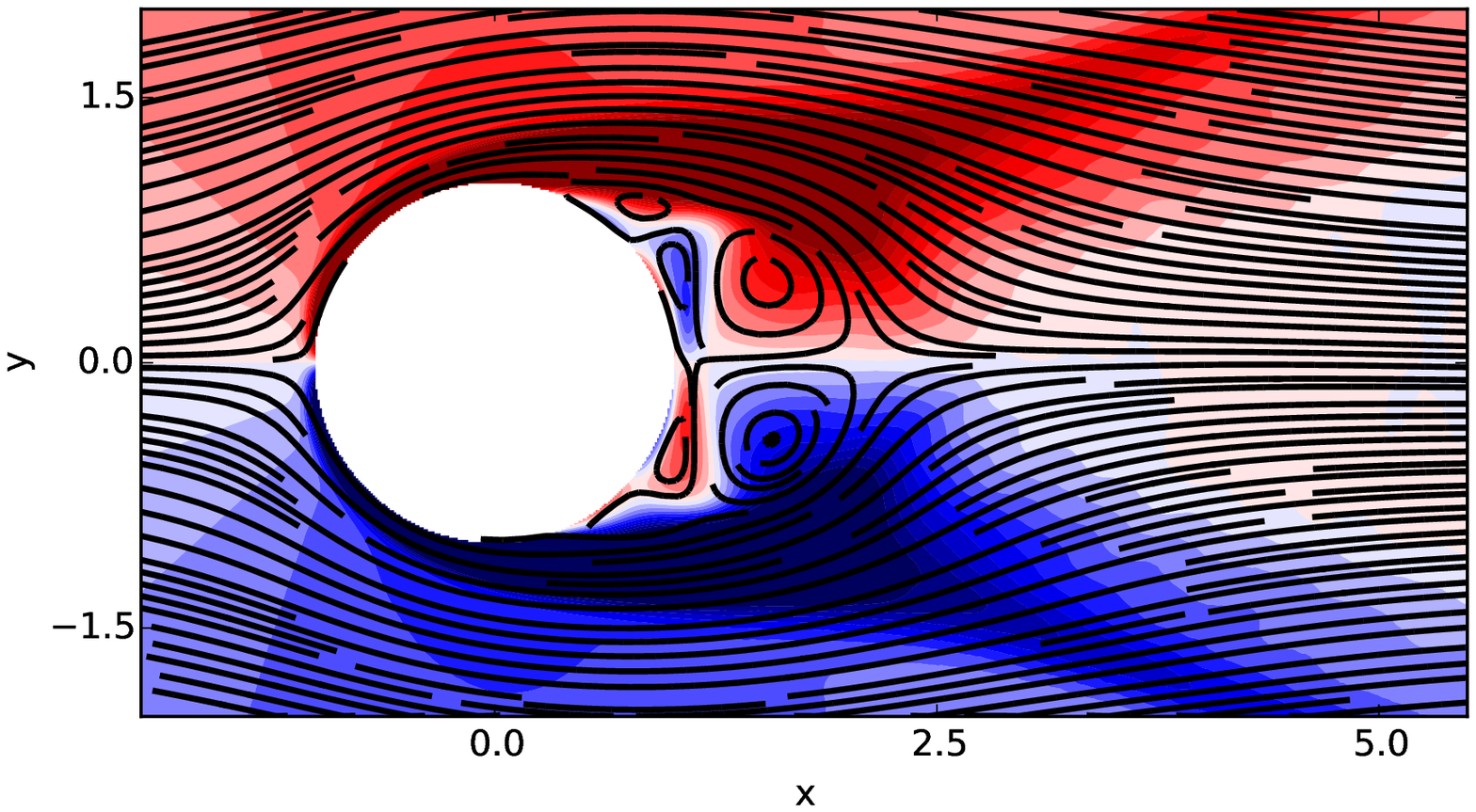}
\label{sfig-snocont}
}\\
\subfigure[]{
\includegraphics[width=.65\linewidth]{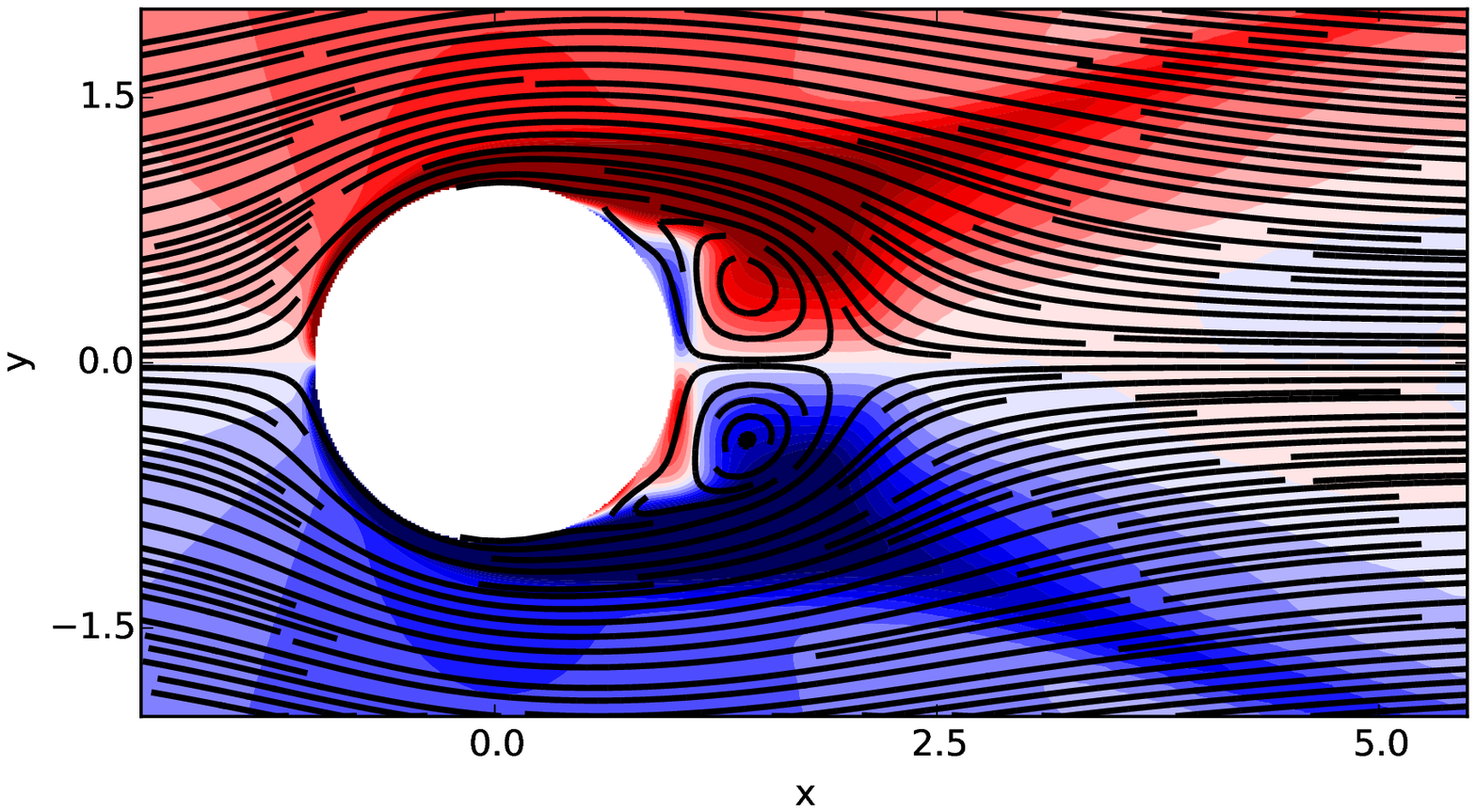}
\label{sfig-scont}
}
\vspace{-1em}
\caption{\small Vorticity field and streamlines for the mean field. (a): for the \NULLtext{} command. (b): for the present identified control strategy. Note that only a part of the computational domain is plotted. Colors encode the sign of the vorticity.}
\label{fig-streamlines}
\end{figure}

\subsubsection{Measurements with noise} \label{sssec-noise}
\revision{To investigate the robustness of our control strategy, the angle of the incident flow is made to vary randomly between $-20$ and $20$ degrees around its nominal value with a uniform probability distribution and a smooth time-evolution, see Fig.~\ref{sfig-angle} for a typical realization.} This mimics a typical class of perturbations to the system at hand and allows the determination of the robustness of the control strategy.

As in the noiseless case considered in Sec.~\ref{sssec-nonoise} above, the transition probability, rewards and Q-factors matrices are all sparse (not shown for sake of brevity).
The performance indicator $\effic$ of the derived policy is depicted in Fig.~\ref{sfig-dist_noise}. In contrast with the noiseless case, $\effic$ is here strongly dependent on the effort in learning.
At early stages of the learning process, the influence of the noise is prominent and the dynamics are not properly captured, which ultimately leads to poor control strategies.

Similarly as in Sec.~\ref{sssec-nonoise}, the values $\left\{\expvalue_i\right\}_{i \in \indS}$, computed from the identified strategy, are larger, on average, than the values from the \oracletext{} policy, see Tab.~\ref{tab-noise_values}.
\revision{This result indicates that the present approach is able to uncover an efficient control strategy, here achieving a significantly lower cost than an optimal time-invariant control strategy, even in a noisy environment.}
\begin{table}[h]
\begin{center}
\revision{
\begin{tabular}{l||c|c|c}
Policy & \NULLtext{} & \oracletext{} & Present strategy \\
\hline
Average value $\left<\expvalue\right>$ & -6.8 &  3.0 & 8.0
\end{tabular}
\caption{Average expected value $\left<\expvalue\right> := \Espoperator_{i \in \indS}\left[\expvalue_i\right]$ associated with the \NULLtext{}, \oracletext{} and present strategy policies. The learning effort for the rewards (resp. the Q-factors) was $10000$.}
\label{tab-noise_values}
}
\end{center}
\end{table}
The resulting control command is plotted in Fig.~\ref{sfig-ncom} and is seen to exhibit an oscillatory behavior. The associated drag coefficient of the cylinder flow is plotted in Fig.~\ref{sfig-ndrag} for the present approach as well the \NULLtext{} and the \oracletext{} strategy. Again, the identified control is seen to perform well and resembles that given by \oracletext{}.

\begin{figure}[t]
\center
\subfigure[]{
\includegraphics[height=.23\linewidth]{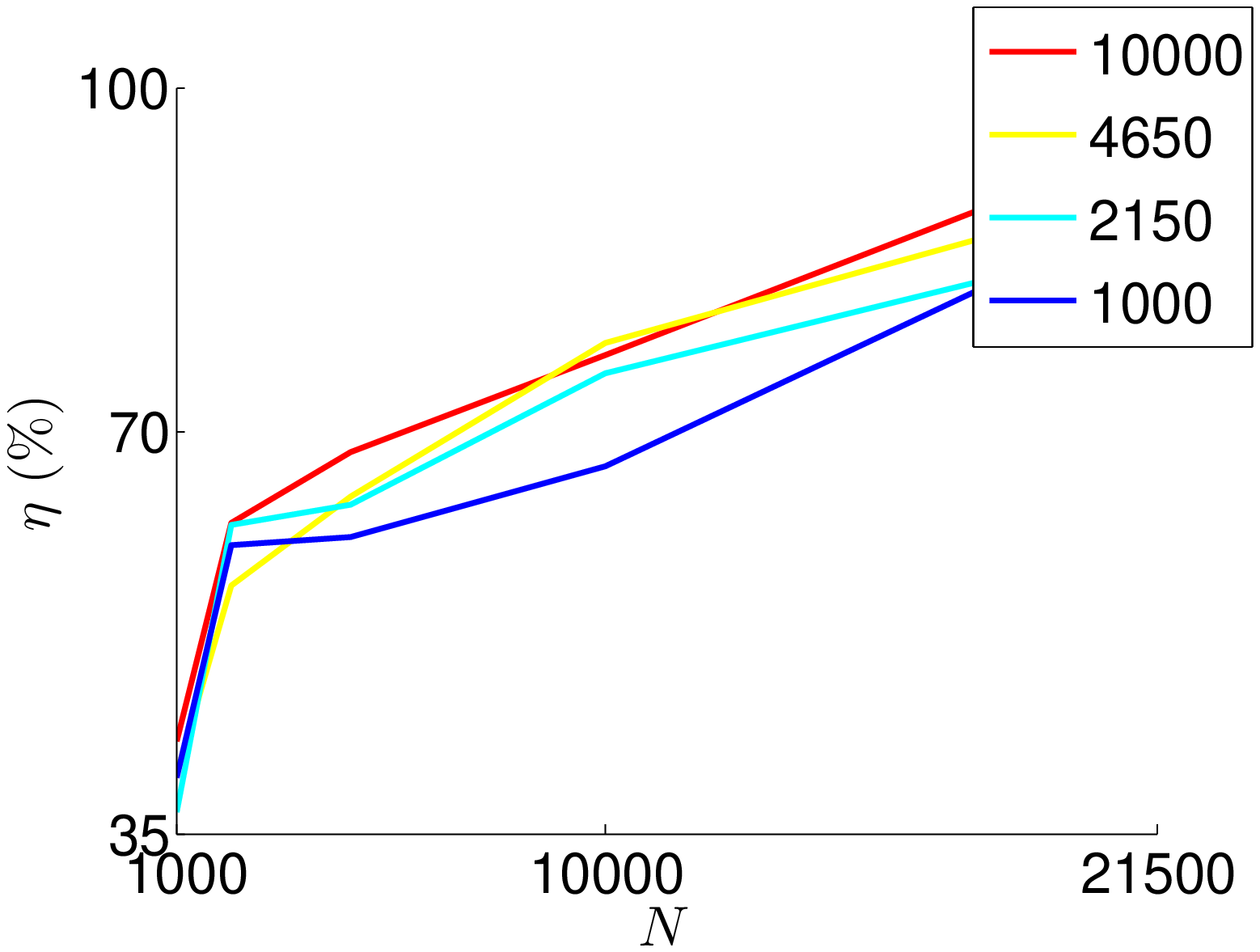}
\label{sfig-dist_noise}
}
\subfigure[]{
\includegraphics[height=.23\linewidth]{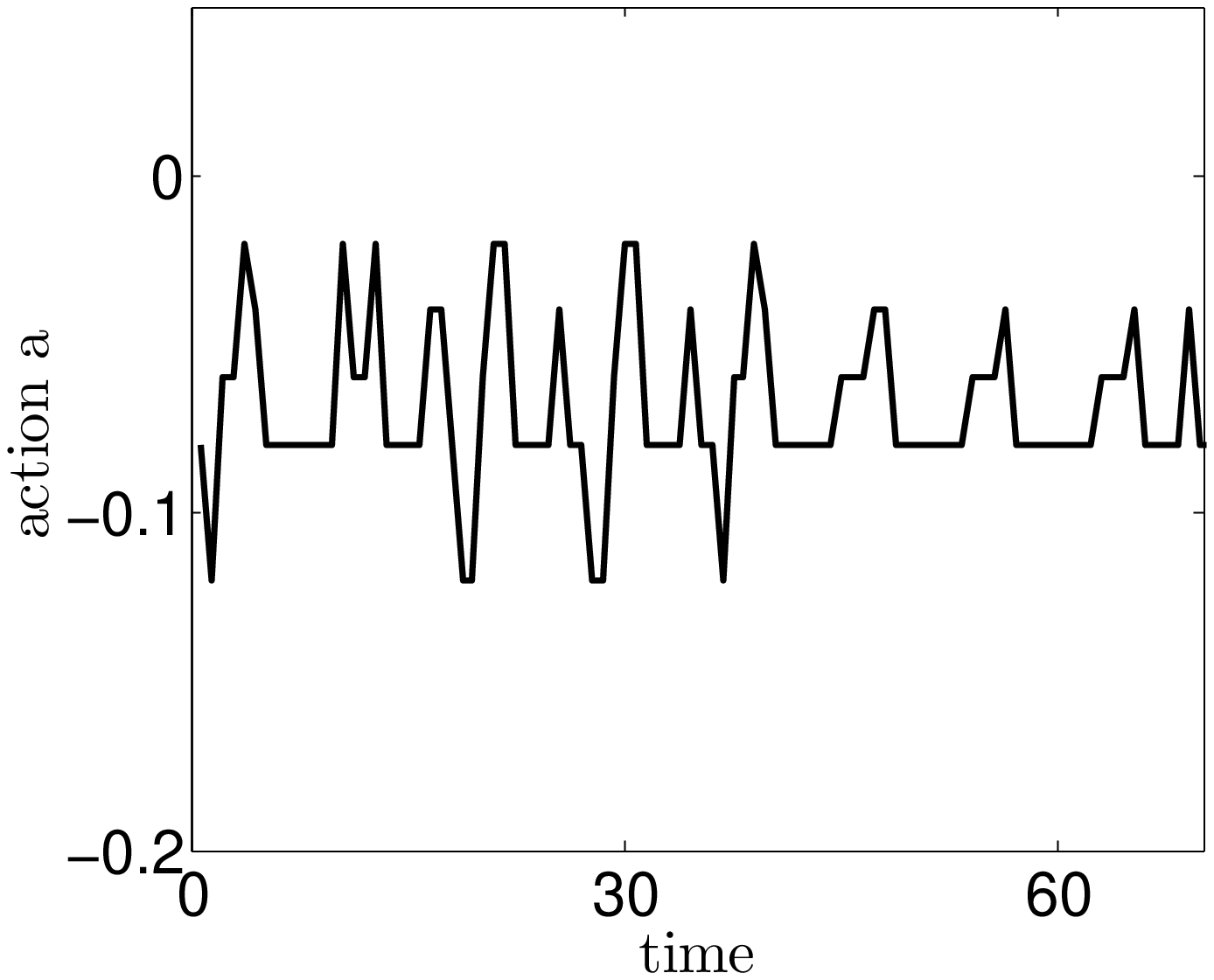}
\label{sfig-ncom}
}
\subfigure[]{
\includegraphics[height=.23\linewidth,width=.30\linewidth]{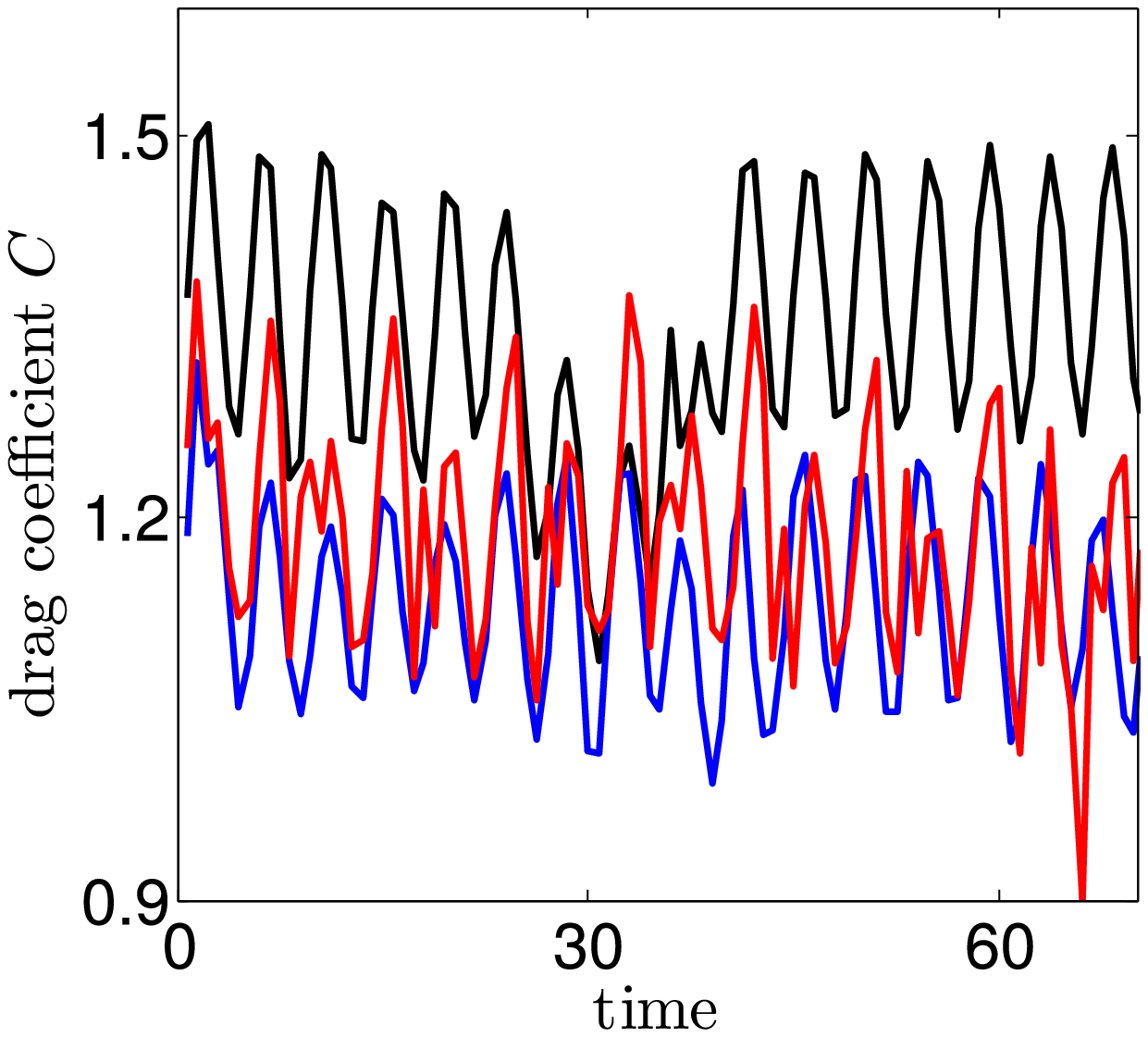}
\label{sfig-ndrag}
}
\vspace{-1em}
\caption{\small Illustration of the control policy.
(a): Performance indicator with respect to the learning effort of the transition rewards matrix $\reward$. Colors (from blue to red) encode different efforts in learning $Q$. (b): Actuation as a function of time, $\action(t)$. (c): Time-evolution of the drag coefficient under three different control policies: zero command (\NULLtext{}, black), \oracletext{} strategy (blue), and the \revision{optimal policy from the} present approach (red). }
\label{fig-ndrag}
\end{figure}

\section{Concluding remarks}
\label{sec-conclu}

This work has presented an experiment-oriented control strategy which does not require any prior knowledge of the physical system to be controlled nor significant computational resources. This strategy allows the learning of a control policy from scarce and point sensors with very limited information on the system at hand. From the sensors' streaming data, a phase space is built using hash functions as a kernel the measurements are convoluted with in real-time. This allows the derivation of a discrete, low-dimensional, space in which the dynamics of the system are approximated. Ensemble-averaged rewards associated with transitions from one discrete state to another are estimated during an online learning sequence. They are directly related to the control objective and tend to sort state transitions based on their impact on the control cost function. A reinforcement learning algorithm is then used to derive the optimal control policy, promoting transitions associated with good rewards.

The resulting method is compliant with actual configurations where instrumentation is limited and spatially-constrained. Owing to the use of kernel hash functions and effective state-aggregation in a discrete phase space, the method runs in real-time and allows closed-loop control. Its discrete and ensemble-averaged nature also brings intrinsic robustness against perturbations in the flow.

This approach has been illustrated on two test cases. The control of a Lorenz system is achieved, by measuring only one component. To mimic a realistic scenario, the actuation is done on a different component. The method is also illustrated by the two-dimensional flow around a circular cylinder. Measurements were provided by a single wall-mounted pressure sensor and actuation was achieved by blowing or suction of fluid at the cylinder surface. The drag coefficient was significantly reduced, reaching essentially the same performance as the control policy given by the \oracletext{} strategy.

More generally, the method presented in this work is readily applicable to physical systems with a causal link between the actuators and the cost functional as evaluated from the sensors. It does not rely on a prior model and instead learns directly from observing the system under stimulation by the actuators, hence being suitable for practical configurations. \revision{An identified limitation of the proposed approach is the number of keys one needs to consider if the relevant dynamics of the system is very rich (large $\Nkey$) and/or a very fine control law is required (large $\Na$). In this situation, the number of entries of the $Q$ matrix grows and hence possibly requires more data for learning. Approximation techniques have however been used to alleviate this limitation \cite{Alex_etal_MIT}.}

Current efforts concern the experimental control of the turbulent flow over an open cavity using the present approach and will be the subject of a subsequent publication. Further developments focus on the convolution kernels, an improved evaluation of suitable rewards and milder assumptions for the reinforcement learning of the control policy.

\bibliographystyle{unsrt}
\bibliography{small_biblio.bib}

\begin{thebibliography}{10}

\bibitem{Gerhard2003}
J.~Gerhard, M.~Pastoor, R.~King, B.R. Noack, A.~Dillmann, M.~Morzynski, and
  G.~Tadmor.
\newblock Model-based control of vortex shedding using low-dimensional galerkin
  models.
\newblock {\em AIAA J.}, 4262(2003):115--173, 2003.

\bibitem{Bergmann2008}
M.~Bergmann and L.~Cordier.
\newblock Optimal control of the cylinder wake in the laminar regime by
  trust-region methods and pod reduced-order models.
\newblock {\em J. Comp. Phys.}, 227(16):7813--7840, 2008.

\bibitem{Ma2011}
Z.~Ma, S.~Ahuja, and C.W. Rowley.
\newblock Reduced-order models for control of fluids using the eigensystem
  realization algorithm.
\newblock {\em Theo. Comp. Fluid Dyn.}, 25(1-4):233--247, 2011.

\bibitem{Joe2011}
W.T. Joe, T.~Colonius, and D.G. MacMynowski.
\newblock Feedback control of vortex shedding from an inclined flat plate.
\newblock {\em Theo. Comp. Fluid Dyn.}, 25(1-4):221--232, 2011.

\bibitem{Mathelin2012}
L.~Mathelin, L.~Pastur, and O.~{Le Ma{\^\i}tre}.
\newblock A compressed-sensing approach for closed-loop optimal control of
  nonlinear systems.
\newblock {\em Theo. Comp. Fluid Dyn.}, 26(1-4):319--337, 2012.

\bibitem{Cordier2013}
L.~Cordier, B.R. Noack, G.~Tissot, G.~Lehnasch, J.~Delville, M.~Balajewicz,
  G.~Daviller, and R.K. Niven.
\newblock Identification strategies for model-based control.
\newblock {\em Exp. Fluids}, 54(8):1--21, 2013.

\bibitem{Lee1997}
C.~Lee, J.~Kim, D.~Babcock, and R.~Goodman.
\newblock Application of neural networks to turbulence control for drag
  reduction.
\newblock {\em Phys. Fluids}, 9(6):1740--1747, 1997.

\bibitem{Kegerise2007}
M.A. Kegerise, R.H. Cambell, and L.N. Cattafesta.
\newblock Real time feedback control of flow-induced cavity tones - part 2:
  Adaptive control.
\newblock {\em J. Sound Vib.}, 307:924--940, 2007.

\bibitem{Huang_Kim_2008}
S.-C. Huang and J.~Kim.
\newblock Control and system identification of a separated flow.
\newblock {\em Phys. Fluids}, 20(10):101509, 2008.

\bibitem{Herve2012}
A.~Herv{\'e}, D.~Sipp, P.J. Schmid, and M~Samuelides.
\newblock A physics-based approach to flow control using system identification.
\newblock {\em J. Fluid Mech.}, 702:26--58, 2012.

\bibitem{Gautier2015}
N.~Gautier, {J.-L.} Aider, T.~Duriez, B.R. Noack, M.~Segond, and M.W. Abel.
\newblock Closed-loop separation control using machine learning.
\newblock {\em J. Fluid Mech.}, 770:442--457, 2015.

\bibitem{Brunton2015}
S.~Brunton and B.~Noack.
\newblock Closed-loop turbulence control: Progress and challenges.
\newblock {\em App. Mech. Rev.}, 67(5):050801, 2015.

\bibitem{Slaney2008}
M.~Slaney and M.~Casey.
\newblock Locality-sensitive hashing for finding nearest neighbors [lecture
  notes].
\newblock {\em IEEE Signal Process. Mag.}, 25(2):128--131, 2008.

\bibitem{Kaiser2014}
E.~Kaiser, B.R. Noack, L.~Cordier, A.~Spohn, M.~Segond, M.~Abel, G.~Daviller,
  J.~Östh, S.~Krajnović, and R.K. Niven.
\newblock Cluster-based reduced-order modelling of a mixing layer.
\newblock {\em J. Fluid Mech.}, 754:365--414, 9 2014.

\bibitem{Mandl1974}
P.~Mandl.
\newblock Estimation and control in markov chains.
\newblock {\em Adv. App. Probab.}, pages 40--60, 1974.

\bibitem{Watkins1992}
C.~Watkins and P.~Dayan.
\newblock Q-learning.
\newblock {\em Mach. Learn.}, 8(3-4):279--292, 1992.

\bibitem{Gosavi2011}
A.~Gosavi.
\newblock Target-sensitive control of markov and semi-markov processes.
\newblock {\em Int. J. Control Autom.}, 9(5):941--951, 2011.

\bibitem{Lin1999}
C.T. Lin and C.P. Jou.
\newblock Controlling chaos by ga-based reinforcement learning neural network.
\newblock {\em IEEE T. Neural Networ.}, 10(4):846--859, 1999.

\bibitem{Gadaleta1999}
S.~Gadaleta and G.~Dangelmayr.
\newblock Optimal chaos control through reinforcement learning.
\newblock {\em Chaos}, 9(3):775--788, 1999.

\bibitem{Grassberger1983}
P.~Grassberger and I.~Procaccia.
\newblock Measuring the strangeness of strange attractors.
\newblock {\em Physica D}, 9:189--208, 1983.

\bibitem{Takens1981}
F.~Takens, D.A. Rand, and L.S. Young.
\newblock Dynamical systems and turbulence.
\newblock {\em Lect. Notes Math.}, 898(9):366, 1981.

\bibitem{Carter1977}
J.L. Carter and M.N. Wegman.
\newblock Universal classes of hash functions.
\newblock In {\em Proc. 9th Ann. ACM Theor. Comp.}, pages 106--112. ACM, 1977.

\bibitem{Andoni2006}
A.~Andoni and P.~Indyk.
\newblock Near-optimal hashing algorithms for approximate nearest neighbor in
  high dimensions.
\newblock In {\em Proc. 47th Ann. IEEE Found. Comp. Sci.}, pages 459--468.
  IEEE, 2006.

\bibitem{Johnson1984}
W.~Johnson and J.~Lindenstrauss.
\newblock Extensions of lipschitz mappings into a hilbert space.
\newblock {\em Contemp. Math.}, 26:189--206, 1984.

\bibitem{Novikov1989}
E.A. Novikov.
\newblock Two-particle description of turbulence, markov property, and
  intermittency.
\newblock {\em Phys. Fluids}, 1(2):326--330, 1989.

\bibitem{Renner2001}
C.~Renner, J.~Peinke, and R.~Friedrich.
\newblock Experimental indications for markov properties of small-scale
  turbulence.
\newblock {\em J. Fluid Mech.}, 433:383--409, 2001.

\bibitem{Bellman1952}
R.~Bellman.
\newblock On the theory of dynamic programming.
\newblock {\em P. Natl. Acad Sci. USA}, 38(8):716, 1952.

\bibitem{Powell2007}
W.~Powell.
\newblock {\em Approximate Dynamic Programming: Solving the curses of
  dimensionality}, volume 703.
\newblock John Wiley \& Sons, 2007.

\bibitem{Lewis2009}
F.~Lewis and D.~Vrabie.
\newblock Reinforcement learning and adaptive dynamic programming for feedback
  control.
\newblock {\em Circuits Syst. Mag., IEEE}, 9(3):32--50, 2009.

\bibitem{Lorenz1963}
E.N. Lorenz.
\newblock Deterministic nonperiodic flow.
\newblock {\em J. Atmos. Sci.}, 20(2):130--141, 1963.

\bibitem{Lemaitre2003}
O.P. {Le Ma\^{\i}tre}, R.H. Scanlan, and O.M. Knio.
\newblock {Estimation of the flutter derivatives of an NACA airfoil by means of
  Navier–Stokes simulation}.
\newblock {\em J. Fluids Struct.}, 17(1):1--28, 2003.

\bibitem{Lusseyran2008}
F.~Lusseyran, L.R. Pastur, and C.~Letellier.
\newblock Dynamical analysis of an intermittency in an open cavity flow.
\newblock {\em Phys. Fluids}, 20(11):114101, 2008.

\bibitem{Alex_etal_MIT}
A.A. Gorodetsky, S.~Karaman, and Y.M. Marzouk.
\newblock Efficient high-dimensional stochastic optimal motion control using
  tensor-train decomposition.
\newblock In {\em Robotics: Science and Systems XI, Sapienza University of
  Rome, Italy, July 13-17}, 2015.

\end{thebibliography}

\end{document}